\documentclass[journal]{IEEEtran}
\usepackage[noend]{algpseudocode}
\usepackage{algorithmicx,algorithm}
\usepackage{amsmath}
\usepackage{graphicx}
\usepackage[colorlinks,linkcolor=blue]{hyperref}
\usepackage{graphicx}
\usepackage{amssymb}
\usepackage{booktabs}
\usepackage{amssymb}
\usepackage{indentfirst}
\usepackage[misc]{ifsym}
\usepackage{amsmath}
\usepackage{multirow}
\usepackage{booktabs}
\usepackage{tikz,xcolor,hyperref}
\usepackage{bbding}
\usepackage{underscore}

\usepackage{amssymb}
\usepackage{amssymb}
\usepackage[misc]{ifsym} 
\usepackage{multirow}
\usepackage{amssymb}
\usepackage{amsmath}
\usepackage{amssymb}
\usepackage{color}
\usepackage{amssymb}
\usepackage{indentfirst}
\usepackage[misc]{ifsym}
\usepackage{amsmath}
\usepackage{multirow}
\usepackage{booktabs}
\usepackage{tikz,xcolor,hyperref}
\usepackage{bbding}
\usepackage{array}
\usepackage{colortbl}
\usepackage{amsmath}
\usepackage{bbding}
\usepackage{pifont}
\usepackage{wasysym}
\usepackage{amssymb}
\usepackage{cite}
\makeatletter

\newcommand{\Rmnum}[1]{\expandafter\@slowromancap\romannumeral #1@}
\makeatother

\begin{document}

%
\title{Towards Homogeneous Modality Learning and Multi-Granularity Information Exploration  for Visible-Infrared Person Re-Identification}
%
%
%

\author{Haojie Liu,
	Daoxun Xia,
	Wei Jiang, and Chao Xu,~\IEEEmembership{Senior Member,~IEEE}
	\thanks{\textit{Corresponding author: Wei Jiang, e-mail: jiangwei_zju@zju.edu.cn}}
\thanks{Haojie Liu is with Huzhou Institute, Zhejiang University, Hangzhou 310027, China (e-mail: liuhaojie@gznu.edu.cn/liuhaojie@stu.xmu.edu.cn).}
\thanks{Wei Jiang and Chao Xu are with the College of Control Science and Engineering, Zhejiang University, Hangzhou 310027, China (e-mail: jiangwei_zju@zju.edu.cn and cxu@zju.edu.cn).}
\thanks{Daoxun Xia is with the School of Big Data and Computer Science and also with Engineering Laboratory for Applied Technology of Big Data in Education, Guizhou Normal University, Guiyang 550025, China (e-mail: dxxia@gznu.edu.cn).}
}

%
%

\markboth{IEEE Transactions on Neural Networks and Learning Systems under review}%
{Shell \MakeLowercase{\textit{et al.}}: Bare Demo of IEEEtran.cls for IEEE Journals}
%



\maketitle

\begin{abstract}
Visible-infrared person re-identification (VI-ReID) is a challenging and essential task, which aims to retrieve a set of person images over visible and infrared camera views. In order to mitigate the impact of large modality discrepancy existing in heterogeneous images, previous methods attempt to apply generative adversarial network (GAN) to generate the modality-consisitent data. However, due to severe color variations between the visible domain and infrared domain, the generated fake cross-modality samples often fail to possess good qualities to fill the modality gap between synthesized scenarios and target real ones, which leads to sub-optimal feature representations. In this work, we address cross-modality matching problem with Aligned Grayscale Modality (AGM), an unified dark-line spectrum that reformulates visible-infrared dual-mode learning as a gray-gray single-mode learning problem. Specifically, we generate the grasycale modality
from the homogeneous visible images. Then, we train a style tranfer model to transfer infrared images into homogeneous grayscale images. In this way, the modality discrepancy is significantly reduced in the image space. In order to reduce the remaining appearance discrepancy, we further introduce a multi-granularity feature extraction network to conduct feature-level alignment. Rather than relying on the global information, we propose to exploit local (head-shoulder) features to assist person Re-ID, which complements each other to form a stronger feature descriptor. Comprehensive experiments implemented on the mainstream evaluation datasets include SYSU-MM01 and RegDB indicate that our method can significantly boost cross-modality retrieval performance against the state of the art methods.
\end{abstract}


%
\IEEEpeerreviewmaketitle

\section{Introduction}\IEEEPARstart{P}{erson} re-identification (Re-ID) , as a fine-grained instance recognition problem, aims to re-identify a query person-of-interest across disjoint camera views \cite{leng2020a, AYe2020, Liu2021a}. Since the surge of deep representation learning, great boosts of Re-ID performance have been witnessed in an idealistic supervised learning testbed: the rank-1 matching rate has reached 96.4\% \cite{Li2019}
on Market1501 dataset, even surpassing human-level recognition rate. 

However, this success relies heavily on an ideal scenario where both probe and gallery images are captured by multiple groups of visible cameras. In real-world scenarios, criminals always appear during the day and commit crimes at night, in which case, visible cameras are incapable of capturing valid appearance information of persons.  To overcome this obstacle, many surveillance cameras automatically toggle their mode from the visible modality to infrared. Accordingly, a new task that associates visible and infrared person images captured by dual-mode cameras for cross-modality image retrieval (VI-ReID) has raised \cite{Wu2017}.

\begin{figure}[tp]
	\begin{center}
		\includegraphics[width=0.99\linewidth]{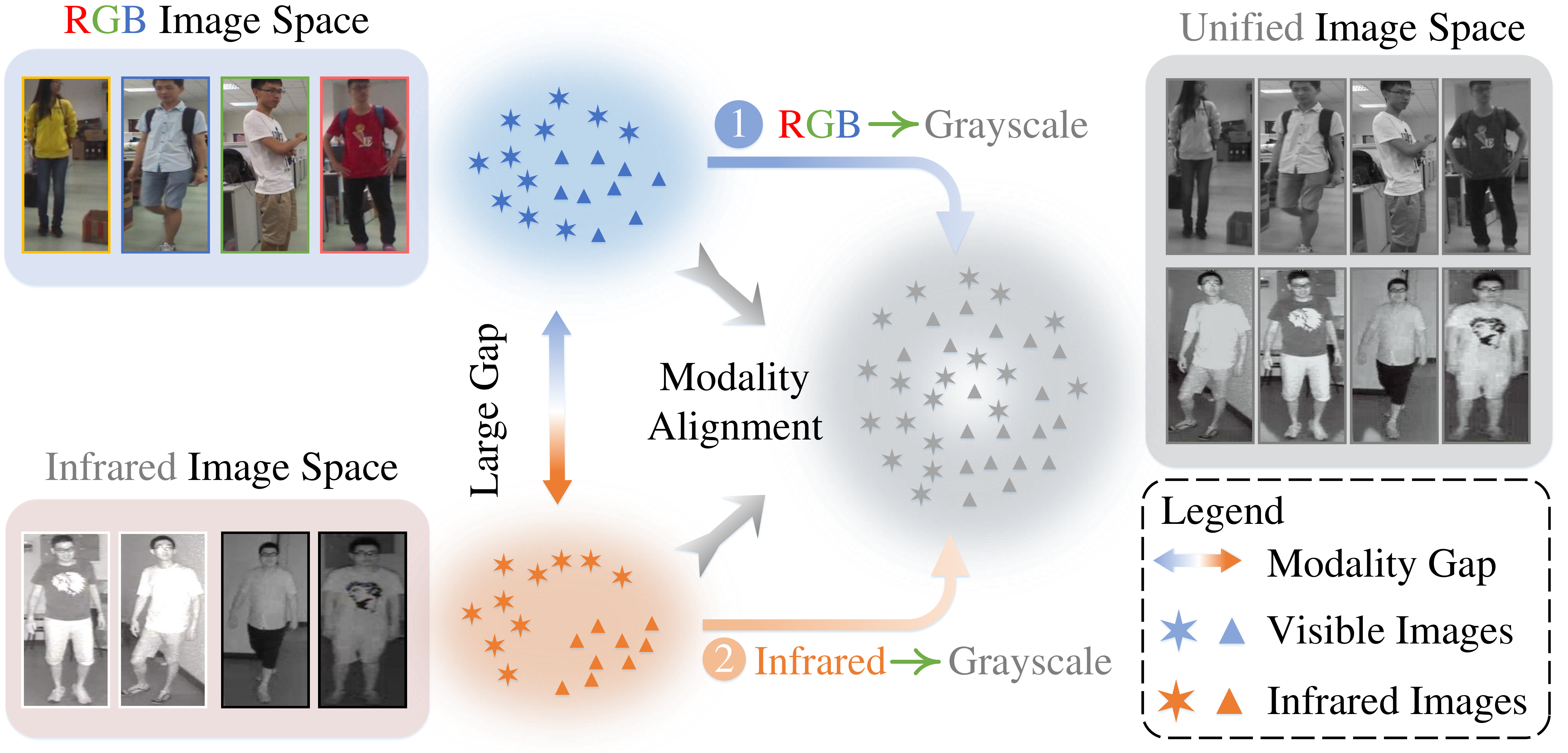}\vspace{-0.4cm}
	\end{center}
	\caption{A high-level overview of homogeneous modality learning strategy. Our method first converts visible images into
		grayscale images, and then uses a style transfer model to transfer infrared images into the grayscale images. In this manner, both modality and luminance gaps are reduced in image-level. Best viewed in color.\vspace{-0.6cm}}
	\label{fig:smalltarget}
\end{figure}\label{sec:introduction}

\begin{figure*}[tp]
	\begin{center}
		\includegraphics[width=0.99\linewidth]{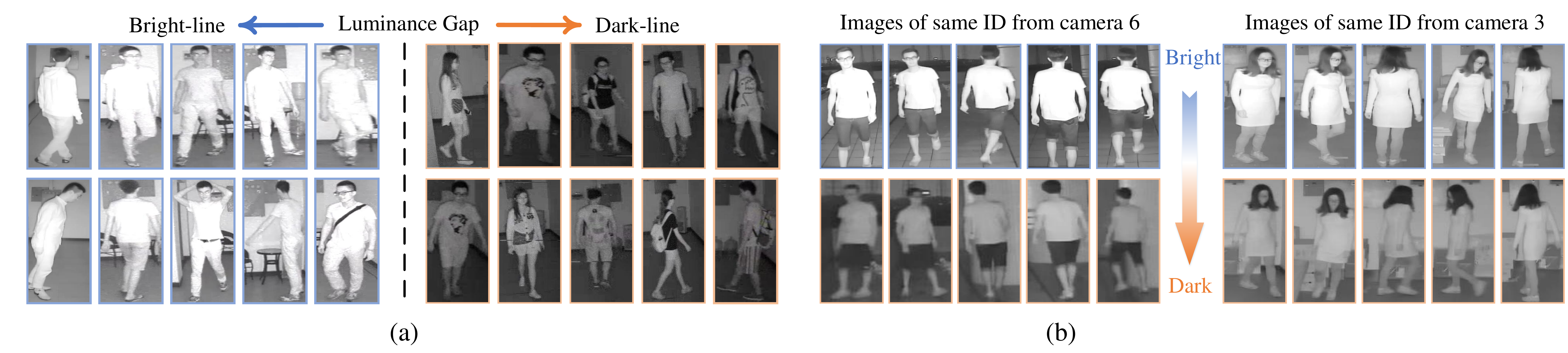}\vspace{-0.4cm}
	\end{center}
	\caption{ Example images from the SYSU-MM01 dataset showing that in addition to the modality discrepancy across visible and infrared modalities, different infrared images also suffer from distinctive luminance variations. \vspace{-0.6cm}}
	\label{fig:smalltarget}
\end{figure*}\label{sec:introduction}

Except for the person's appearance discrepancy involved in single-modality ReID, VI-ReID encounters the additional modality discrepancy resulting from the different imaging processes of spectrum cameras. In an effort to minimize such modality gap, one representative method-of-choice is to embed heterogeneous images into a shared feature space so as to align feature distribution using feature-level constraints \cite{Wu2017, Ye2018, Ye2020, Ye2018a}. However, feature-optimization based model in practice is often constrained in a homogeneous feature space. While for heterogeneous images, it is always a suboptimal problem.
Another line is image synthesis methods \cite{Wang2019, Wang2020, Xia2021, Wang2019a}, which exploit generative adversarial networks (GANs) as a style transformer to generate multi-spectral images. However, due to the insufficient amount of cross-modality paired examples, the generative pipeline often leads to low-fidelity generations (incomplete local structure or unavoidable noise). If we directly use these low-quality synthetic images to train an Re-ID model, a novel gap between the original data and the synthetic data would be introduced to the learning process, thereby undermining the training process.

Above liminations promote us to consider: if there exists one high-fidelity shared image space that the different modality information can be treated equally? In other words, we only need to eliminate person apperance discrepancy in such space, just same as the goal of conventional single-mode Re-ID methods.
Motivated by this train of thought, in this paper, we explore the correlation between
two modalities and formulate a unified spectral to improve the similarity of feature distributions, called Aligned Grayscale Modality (AGM). As shown in Fig. 1, our method is divided into two steps. First, we obtain grayscale images from visible images directly by image graying operation. Second, with generated grayscale images, we train a style transfer model to transfer the style of infrared images into grayscale. In this way, heterogeneous modality data are aggregated into homogeneous modality data.
Comprared to existing GAN-based algorithms, the proposed AGM 1) perfectly persists the discriminative information of orignial images, 2) really and truly actualizes the modality discrepancy elimination in image-level, and 3) is easy to implement without dizzy training strategies.

In addition to fulfil visible-infrared modality alignment, AGM also suppresses the gap of infrared image brightness changes.  Specifically, as shown in Fig. 2(a), the left infrared images present a highlighted appearance, while the right show the low brightness. The same observation can also be seen in Fig. 2(b): the top row presents bight spectrum, yet the bottom row of the same identity presents dark spectrum. We formulate this phenomenon as `luminance gap', which produces terrible influence. In this paper, AGM defuses such luminance gap using CycleGAN, that all infrared images are normalized into homogeneous grayscale images. We call this process as the Grayscale Normalization (GN). Benefiting from the unique grayscale style, the normalized infrared images can perfectly clear up the luminance gap.

Then, to reduce the remaining appearance discrepancy, we propose to leverage the head-shoulder information to assit global features. The head-shoulder area possesses abundant discriminative information, such as hair-style, face and neckline style, that play an important role in inferring the interested target person. In particular, as shown in Fig. 3, we design a two-stream cascade structure to encode both finer-granularity (head-shoulder) and coarser-granularity (global) appearance information. Then, we concatenate two types of features for generating the final person representation and back-propagate the supervised loss to all specific and joint branches. Mutual interaction of head-shoulder and global information can obviously enhance the feature representation ability, however, this behaviour is always conducted as an asynchronous learning scheme in different branches. Therefore, in order to ensure synergistically correlated feature learning at different branches, we also develop a synchronous learning strategy (SLS). It explicitly optimises the underlying complementary advantages across granularities via imposing a closed-loop cross-branch interactive regularisation. Under such balance between individual learning and correlation learning in a closed-loop form, we allow all branches to be learned concurrently in an end-to-end fashion.


To summarize, we make the following contributions:

$\bullet$ we attempt an under-explored but significant research path for addressing cross-modality problem. In particular, we creat a unified middle modality image space to embed the homogeneous modality information, which builds a connection between visible and infrared domains. It is worth recalling that, our middle modality space (AGM) is fully visual, high-fidelity and easily reproduced. We believe AGM has great potential to further boost cross-modality retrieval performnce.



$\bullet$ To further make clear cross-modality matching challenges, we for the first time introduce a new concept
called Luminance Gap. This leads
to Grayscale Normalization (GN), a style-based normalized approach capable of suppressing the luminance varitions of infrared images and further alleviates the modality discrepancy.

$\bullet$ We investigate the multi-granularity feature learning problem and formulate a more robust head-shoulder descriptor to support person Re-ID matching. Head-shoulder part effectively augments person information with discriminative appearance cues to construct high dimensional fusion features, leading to a competitive Re-ID performance.  

$\bullet$ A synchronous learning strategy (SLS) with a well-designed closed-loop interactive regularisation is developed to optimize the underlying complementary advantages of both global and head-shoulder information, that urges the network to obtain more discriminative features for correct classification.

\section{Related Work}

\subsection{Visible-Visible Re-ID Methods.}
Visible-visible person Re-ID studies typically tackle a single-modality case, that is, both query and gallery images are captured by visible cameras. It usually suffers from the large intra-class variations caused by different views \cite{Zhu2020}, poses \cite{Liu2018} and occlusions \cite{Miao2019}. Nowadays, substantial research efforts \cite{Sun2019, Lin2019, Luo2019, Zhang2020, Li2019, Wang2018, Varior2016, Fan2019, Hermans2017} have been constructed to extract discriminative features or learn effective distance metrics. For a instance, the work of \cite{Lin2019} exploits attributes as complementary information to help recognize the target person. Self-attention based methods \cite{Luo2019, Zhang2020} incorporate attention techniques to let the network concentrate on discriminative regions. Part-based approaches \cite{Sun2019, Li2019, Wang2018} treat person Re-ID as a partial feature learning task, dividing person images into multiple horizontal strips and applying independent classifiers to suprvise each local strips. Other methods are based on metric learning, focusing on desgining proper loss functions for optimizing feature distances between different samples, like the contrastive loss \cite{Varior2016}, sphere loss \cite{Fan2019} and triplet loss \cite{Hermans2017}. The overwhelming majority of techniques in these literature have achieved considerable success in visible-to-visible matching, while they are ill-suited for cross-modality image retrieval in poor lighting environments \cite{Wu2017}, limiting
applicability in practical 24-hour surveillance situations.

\subsection{Visible-Infrared Re-ID Methods.}
Visible-infrared person Re-ID task is proposed to achieve 24-hour continuous surveillance. In addition to conventional appearance discrepancy, it also suffers from the modality discrepancy originating from different wavelength ranges of spectrum cameras \cite{Wu2017}. To handle such cross-modality discrepancies, early works try to learn a modality-sharable feature representation using feature-level constraints \cite{Wu2021, Ye2018, Ye2018a, Ye2020, Liu2021}. They design novel classification and/or triplet losses for pointing at optimizing cross-modality samples. Specifically, \cite{Ye2020} uses modality-sharable and modality-specific classifiers to learn identity information in the classifier level and introduce a collaborative ensemble
learning scheme to collaboratively optimize the feature learning with multiple classifiers. \cite{Ye2018} propose a bi-directional top-ranking loss, which samples positive and negative pairs from different modalities and optimizes such cross-modality triplets with a bi-directional interactive iteration manner. More recently, some other works adopt adversarial training strategies to reduce the cross-modality distribution divergence in image-level \cite{Wang2019, Wang2020, Xia2021, Wang2019a, Zhong2021, Zhang2021}. For a instance, they transfer stylistic properties of visible images to their infrared counterpart, with an identity-preserving constraint \cite{Wang2019a,Wang2019} or cycle consistency \cite{ Wang2020, Xia2021}. However, due to the lack of paired cross-modality training data, GAN-based methods always involve much randomness, which may lead to identity inconsistency during the complicated adversarial training proces \cite{Xia2021, Wang2019a}. In contrast, our method proposes to exploit aligned grayscale modality space (AGM) to reduce the cross-modality distribution divergence in image-level. It is no longer the pattern of transfering A to B or B to A, but projecting A and B to C, where the space of C treats the different modality information equally. 

\begin{figure*}[tp]
	\begin{center}
		\includegraphics[width=0.99\linewidth]{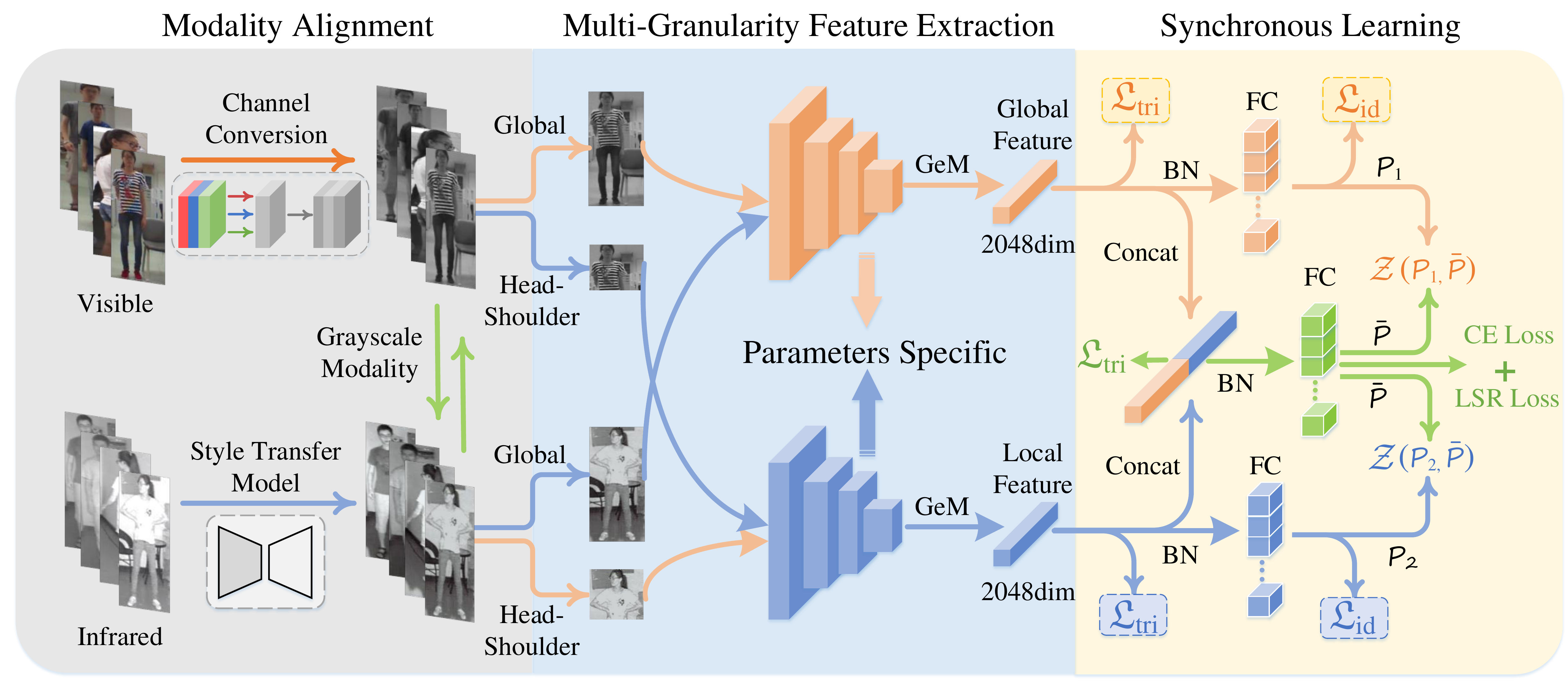}\vspace{-0.6cm}
	\end{center}
	\caption{The proposed framework for VI-ReID which contains modality alignment module, multi-granularity feature extraction module and multi-branch synchronous learning module. The grayscale features generated via modality alignment module are directly exploited for modality-sharable feature learning. For multi-granularity feature extraction, both gobal and head-shoulder appearance information are encoded by two branches for producing the specific features, and the multi-granularity fusion branch produces the final Re-ID features for learning the consensus on identity classes across two sub branches. The training of each branch is supervised by the same identity class label and triplet constraints concurrently.\vspace{-0.4cm}}
	\label{fig:smalltarget}
\end{figure*}\label{sec:introduction}

\subsection{Finer-granularity Information.}
Finer-granularity information, such as clothing, hair style, etc., produce abundant discriminative feature representations for contributing the person Re-ID, especially when color information is entirely uninformative in visible-infrared and gray-gray matching problem. However, as we know, it has been rarely explored and remains an open issue. The literature \cite{Sun2019} is the pioneering work to attempt to improve Re-ID performance with part features. It generates part-level features by partitioning
the convolutional tensor and calculates the cross-entropy loss for
every achieved part-level column vector. To make the model make robust in crowded conditions, another work \cite{Li2018} focuses solely on head-shoulder part instead of the whole body for person Re-ID. It splits head-shoulder images into groups by pose pairs and trains similarity classifier for each. Due to different pose features are ambiguous for naive classifiers, an ensemble conditional probability is leaned for excavating relationship among multiple poses. Inspired by the basic idea of head-shoulder information \cite{Li2018}, we present a two-stream cascade structure to simultaneously encode global and head-shoulder part features for gray-gray Re-ID problem, revolutionizing the method of local feature assisting global feature in the existing literature.

\section{Proposed method.}
In this section, we present the structure of the proposed aligned grayscale modality (AGM) learning model, which is aimed at learning robust modality-invariant feature representations for visible-infrared person Re-ID. As shown in Fig. 3, we first formulate a unified middle modality space to overcome modality discrepancy. Then, we introduce our two-stream cascade network for learning high-level semantic features of both coarser-granularity global and finer-granularity head-shoulder inputs. Finally,  to discover and capture correlated complementary combination between the global and head-shoulder features, we supervise each branches with the same identity class label and further introduce a synchronous learning strategy (SLS) to regulate iterative learning behaviour together.

\subsection{Aligned Grayscale Modality Generation Module.}
\subsubsection{Visible to Grayscale Image Transformation}
Like the visible image, the description of grayscale image still reflects the distribution and characteristics of global and local chromaticitys of the whole image, simultaneously well approximating the style of infrared image. Therefore, when conventional appearance cues such as colors and textures get unreliable for the person matching, grayscale image is the best choice to replace visible images for feature learning. Given a visible image $x_v^i$ with three channels $\mathcal{R,G,B}$, we read the each pixel point $\mathcal{R}(x)$, $\mathcal{G}(x)$ and $\mathcal{B}(x)$ values of the visible image $x_v^i$ in turn. The corresponding grayscale pixel point $\mathcal{G}(x)$ then can be calculated as:
\begin{equation}
\mathcal G(x)= \alpha_1 \mathcal R(x) + \alpha_2 \mathcal G(x) + \alpha_3 \mathcal B(x),
\end{equation}
where $\alpha_1$, $\alpha_2$ and $\alpha_3$ are set to $0.299$, $0.587$ and $0.114$, respectively. The generated grayscale value $\mathcal G(x)$ is averagely distributed to each channels ($\mathcal{R}$, $\mathcal{G}$, $\mathcal{B}$) of the original visible image, so that all grayscale images still have three channels that can be fed into the deep model. 

\subsubsection{Infrared to Grayscale Image Transformation}
In this section, we present the process of infrared image to grayscale image transformation, which is also called gray noramlization (GN). This is achieved by cycle-consistent adversarial networks (CycleGAN) \cite{Zhu2017}. GN 
can significantly address two following problems for VI-ReID task: (1) smoothing the luminance disparities of different infrared images, and (2) further alleviating the slight modality discrepancy between infrared and grayscale domains. 

\textbf{Formulation.} We define two sets of training images $X^g$ and $X^t$, collected from two different domains $\mathcal{A}$ (grayscale) and $\mathcal{B}$ (infrared), where $X^g \in \mathcal{A}$ and $X^t \in \mathcal{B}$. Specifically, $X^g$ contains images from the grayscale modality (denoted by $X^g = \{x_i^g\}_{i=1}^\mathcal{M}$) and $X^t$ contains images from the infrared (thermal) modality (denoted by $X^t = \{x_i^t\}_{i=1}^\mathcal{N}$). $\mathcal{M}$ and $\mathcal{N}$ represent the number of grayscale and infrared images in their training set respectively. Additionly, we also denote the sample distribution of grayscale and infrared domains as: $x^g \sim p_{data}(x^g)$ and $x^t \sim p_{data}(x^t)$. Two mapping generators are defined as: $G: \mathcal{A} \rightarrow \mathcal{B}$, $F: \mathcal{B} \rightarrow \mathcal{A}$, and two adversarial discriminators are defined as: $D_{\mathcal{A}}$, $D_{\mathcal{B}}$. Our goal is to learn a mapping function such that the generated distribution of images $G(X^t)$ is indistinguishable from the target distribution $p_{data}(x^g)$.

\textbf{Adversarial Loss.} We apply adversarial losses \cite{Goodfellow2014} to both mapping functions by using the cross-reconstructed images with different modalities. In the case of grayscale modality, the discriminator $D(\mathcal{A})$ distinguishes the real image $x^g$ and the generated fake image $G(x^t)$. Similarly, in the case of infrared modality, the discriminator $D(\mathcal{B})$ distinguishes the real image $x^t$ and the generated fake image $G(x^g)$. Here, the generator $G(.)$ ($F(.)$) try to synthesize more realistic images that look similar to images from domain $\mathcal B$ ($\mathcal A$). Formally, adversarial losses involve finding a Nash equilibrium to the following two player min-max problem:

\begin{equation} 
\begin{split}
\mathcal{L}^{adv}_{x^g\rightarrow x^t} (G, D_{\mathcal{B}}) = &\mathbb E_{x^t \sim p_{data}(x^t)}[logD_{\mathcal{B}}(x^t)] \\
+&\mathbb E_{x^v \sim p_{data}(x^v)}[1-logD_{\mathcal{B}}(G(x^g))],
\end{split}
\end{equation}

\begin{equation} 
\begin{split}
\mathcal{L}^{adv}_{x^t\rightarrow x^g} (F, D_{\mathcal{A}}) = &\mathbb E_{x^g \sim p_{data}(x^g)}[logD_{\mathcal{A}}(x^g)] \\
+&\mathbb E_{x^t \sim p_{data}(x^t)}[1-logD_{\mathcal{A}}(G(x^t))],
\end{split}
\end{equation}
where $x^g\rightarrow x^t$ ($x^t\rightarrow x^g$) means mapping grayscale (infrared) domain to infrared (grayscale) domain, respectively. The discriminative networks ($D_{\mathcal{A}}$ and $D_{\mathcal{B}}$) are trained in an alternating optimization alongside with the generators $G, F$. Especially, the parameters of the discriminator are updated when the parameters of the
generator are fixed.

\begin{figure*}[tp]
	\begin{center}
		\includegraphics[width=0.99\linewidth]{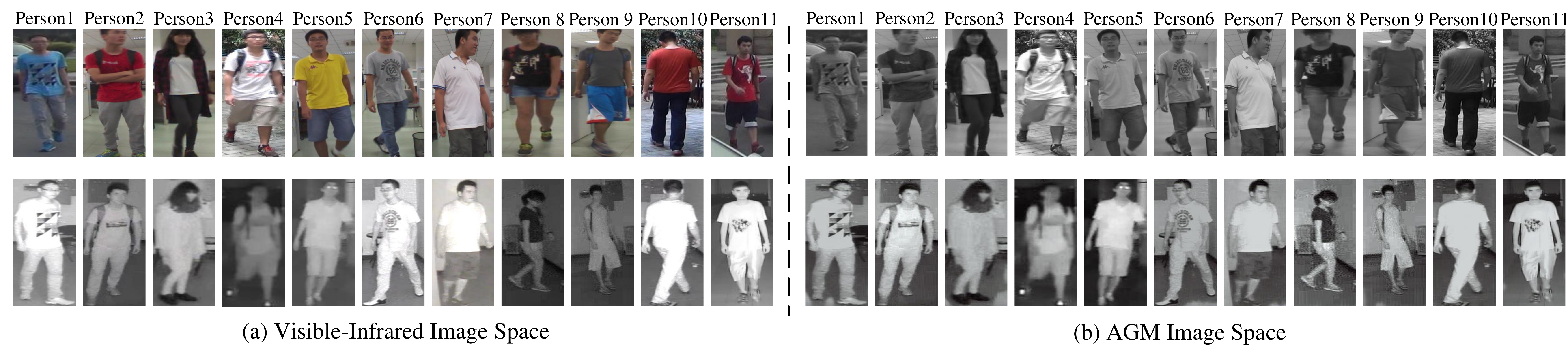}\vspace{-0.6cm}
	\end{center}
	\caption{Contrast visualization between the raw modality images (a) and the aligned grayscale modality images (b). From left to right, the two images of a column share the same identity. It can be obviously observed that the raw visible-infrared image space suffers from both large modality and luminance gap. In contrast, our proposed AGM image space perfectly construct a modality discrepancy free space for cross-modality matching. \vspace{-0.4cm}}
	\label{fig:smalltarget}
\end{figure*}\label{sec:introduction}

\textbf{Cycle Consistency Loss.} Adversarial training strategy, in practice, forces generators $G$ and $F$ to produce outputs identically distributed as target domains $\mathcal{A}$ and $\mathcal{B}$. However, for cross-modality image-to-image translation issue, we hope transfered images only change its style to fit the target domain, while the whole semantic information is still retained throughout the conversion process. Thus, inspired by CycleGAN \cite{Zhu2017}, we apply a cycle consistency loss:
\begin{equation} 
\begin{split}
\mathcal{L}_{cyc}(G, F) = &\mathbb E_{x^g \sim p_{data}(x^g )}[||F(G(x^g) )-x^g ||_{1}] \\
+ & \mathbb E_{x^t  \sim p_{data}(x^t)}[||G(F(x^t))-x^t||_{1}],
\end{split}
\end{equation}
where $F(G(x^g))$ and $G(F(x^t))$ are the cycle-reconstructed images, respectively.

\textbf{Identity mapping loss.} Additionally, to encourage mapping to maintain the consistency of input and output colors, we adopt an identity regularization term to assit the generator to be near an identity mapping when using real images of the target domain as input, which is defined as:
\begin{equation} 
\begin{split}
\mathcal{L}_{identity}(G, F) = &\mathbb E_{x^t \sim p_{data}(x^t )}[||G(x^t) -x^t ||_{1}] \\
+ & \mathbb E_{x^g  \sim p_{data}(x^g)}[||F(x^g)-x^g||_{1}].
\end{split}
\end{equation}

These lead to the final objective functions:
\begin{equation} 
\begin{split}
\mathcal{L} (G,F, D_{\mathcal{A}}, D_{\mathcal{B}}) = &\mathcal{L}^{adv}_{x^g\rightarrow x^t} (G, D_{\mathcal{B}}) \\
+&\mathcal{L}^{adv}_{x^t\rightarrow x^g} (F, D_{\mathcal{A}})\\
+& \lambda_1 \mathcal{L}_{cyc}(G, F)\\
+& \lambda_2 \mathcal{L}_{identity}(G, F),
\end{split}
\end{equation}
where $\lambda_1$ and $\lambda_2$ control the relative importance of the two objectives. In Fig. 4, we show more qualitative results of our aligned grayscale modality generation images.

\subsection{Multi-Granularity Feature Extraction Module.}
\subsubsection{Data Extraction of Head-shoulder Area}
We directly collect person head-shoulder data cropped from benchmarks and form them into an independent training set. Specifically, assume that the size of a global training image $x$ is $\mathcal W \times \mathcal H$. We generate corresponding head-shoulder image with retaining the upper third of the original image, the size of which is $\mathcal W \times (\mathcal H/3)$. Take the upper left corner of the global image $x$ as the origin and establish a rectangular coordinate system in pixels, the coordinate of $x$ can be formulated as [0, 0, $\mathcal W$, $\mathcal H$]. Then, we directly crop image according to the coordinate of head-shoulder point [0, 0, $\mathcal W$, $\mathcal H/3$]. The above process is repeated until every global images have its corresponding head-shoulder images.


\subsubsection{Network Structure}
As shown in Fig. 3, our multi-granularity feature extraction framework consists of two learnable branches with independent parameters. The first branch is set as global feature extractor to encode coarser-granularity appearance information, while the second branch undertakes the work of extracting  finer-granularity head-shouler features. For preciseness in presentation, we
denote the global stream feature extraction network as function $\mathcal F_g(.)$ and the head-shoulder stream feature extraction network as as function $\mathcal F_h(.)$. 
Given a global input image $x_i^g (i\in \mathcal{N})$, the global stream feature extraction network outputs a convolutional feature map $F_i^g \in \mathbb{R}^{C\times H_1 \times W_1}$, which meets:
\begin{equation} 
F_i^g = \mathcal F_g(x_i^g;\boldsymbol{\theta}_{\mathcal F_g}),
\end{equation}
where $\mathcal{N}$ is the number of global training images in a mini-batch and $\boldsymbol{\theta}_{\mathcal F_g}$ represents the parameter of the global branch $\mathcal F_g(.)$. $C, H_1$ and $W_1$ denote the channel, height and width dimension of global output feature maps.

Similarly, for a head-shoulder input  image $x_i^h$, the head-shoulder stream feature extraction network also outputs a corresponding convolutional feature map $F_i^h \in \mathbb{R}^{C\times H_2 \times W_2}$,
\begin{equation} 
F_i^h = \mathcal F_h(x_i^h;\boldsymbol{\theta}_{\mathcal F_h}),
\end{equation}
where $\boldsymbol{\theta}_{\mathcal F_h}$ represents the parameter of the head-shoulder branch $\mathcal F_g(.)$. $C, H_2$ and $W_2$ denote the channel, height and width dimension of the head-shoulder output feature maps.

Then, inspire by the work \cite{AYe2020}, generalized mean pooling layer (GeM) is employed on the top of feature extractors to acquire a compact embedding vector in the common space. The extracted embedding vectors ($\mathcal V_i^g$, $\mathcal V_i^h$) from two global and head-shoulder branches are formulated as:
\begin{equation} 
\mathcal V_i^g = \mathrm{GeM}(F_i^g); \mathcal V_i^h = \mathrm{GeM}(F_i^h),
\end{equation}
where $\mathrm{GeM}(.)$ denotes the operator of the generalized mean pooling layer. Finally, we merge these embedding vectors of both the global and head-shoulder branches into a new joint branch to obtain the fused person feature, that is:
\begin{equation} 
\mathcal V_i^{joint} = \mathcal V_i^g \oplus \mathcal V_i^h,
\end{equation}
where $\mathcal V_i^{joint}$ denotes the joint feature and $\oplus$ means concatenate method. Note that $\mathcal V_i^{joint}$ is used as the final representation for
Person Re-ID.

\subsubsection{Common Feature Space Constraints}
\ 

\textbf{Hard Mining Triplet Loss:}
The motivation of the triplet loss \cite{Schroff2015} is to optimize the distance threshold for separating positive and negative objects, making embedding vectors from same classes produce more obvious clustering results in the common feature space.
Here, for three extracted embedding vectors $\mathcal V^{joint}$, $\mathcal V^{g}$, $\mathcal V^{h}$, we adopt a batch hard mining triplet loss \cite{Hermans2017} to optimize the relative distance between positive and negative pairs of themselves simultaneously.

Given a mini-batch of global person embedding vectors $\{\mathcal V^g_i \}_{i=1}^{\mathcal{N}}$, we sample a feature triplet ($\mathcal V^g_a$, $\mathcal V^g_p$, $\mathcal V^g_n$) where the hardest positive point $\mathcal V^g_p$ is from the same class with the anchor point $\mathcal V^g_a$ and the hardest negative point $\mathcal V^g_n$ is from different identities with $\mathcal V^g_a$.
Hard mining triplet loss forces all points belonging to the same class to form a single cluster and pushes other negative samples forward:
\begin{equation}
\begin{split}
\mathcal{L}_{t}^{g}(\boldsymbol{\theta}_{\mathcal F_g}) &= \frac{1}{\mathcal{N}}\sum_{(a,p,n)}[\mathop{max}\limits_{\forall a=p} \mathbf{D}((\mathcal V^g_a), (\mathcal V^g_p))\\
& -\mathop{min}\limits_{\forall a\neq n} \mathbf{D}((\mathcal V^g_a), (\mathcal V^g_n))+\xi ]_+,
\end{split}
\end{equation}
where $\mathbf{D}(.)$ represents the Euclidean Distance between two feature vectors and $[ . ]_+ = \mathbf{max}(x,0)$ represents a hinge loss. For learning multi-granularity fused features, we formulate the hard mining triplet loss for other branches as the following:
\begin{equation}
\begin{split}
\mathcal{L}_{t}^{h}(\boldsymbol{\theta}_{\mathcal F_h}) &= \frac{1}{\mathcal{N}}\sum_{(a,p,n)}[\mathop{max}\limits_{\forall a=p} \mathbf{D}((\mathcal V^h_a), (\mathcal V^h_p))\\
& -\mathop{min}\limits_{\forall a\neq n} \mathbf{D}((\mathcal V^h_a), (\mathcal V^h_n))+\xi ]_+,
\end{split}
\end{equation}
\begin{equation}
\begin{split}
\mathcal{L}_{t}^{joint}(\boldsymbol{\theta}_{\mathcal F_g}, \boldsymbol{\theta}_{\mathcal F_h}) &= \frac{1}{\mathcal{N}}\sum_{(a,p,n)}[\mathop{max}\limits_{\forall a=p} \mathbf{D}((\mathcal V^{joint}_a), (\mathcal V^{joint}_p))\\
& -\mathop{min}\limits_{\forall a\neq n} \mathbf{D}((\mathcal V^{joint}_a), (\mathcal V^{joint}_n))+\xi ]_+.
\end{split}
\end{equation}
Here, $\mathcal{L}_{t}(\boldsymbol{\theta}_{\mathcal F_g})$ and $\mathcal{L}_{t}(\boldsymbol{\theta}_{\mathcal F_h})$ aims to optimize the parameters of global and head-shoulder branches repectively. The joint triplet loss $\mathcal{L}_{t}(\boldsymbol{\theta}_{\mathcal F_g}, \boldsymbol{\theta}_{\mathcal F_h})$ can further fine-tune the
concatenated features for both global and head-shoulder branches.

\textbf{Identity Loss:} The identity loss $\mathcal{L}_{id}$ is a softmax function based cross entropy loss widely used in classification tasks. It utilizes cosine distance to separate the embeded space into different subspaces for optimizing person identity discrimination. Formally, we predict the posterior probability $p(y_i|x^g_i)$ of the global training image $\{x^g_i\}_{i=1}^{\mathcal{N}} $ over the given identity label $y_i$:
\begin{equation} 
p({y_{i}|x_i^g})=\frac{exp(W^T_{y_{i}}\times \mathcal V_i^g)}{\sum_{k=1}^{\mathcal{N}}exp(W^T_{k}\times \mathcal V_i^g)}, k = 1, 2, ..., \mathcal{N},
\end{equation}
where $\mathcal V_i^g$ refers to the embedding feature vector of $x_i^g$ from the global branch. $W_k$ is the weight parameter matrix of the last fully connected layer for $k$ th identity. The global branch model identity training loss is computed as:
\begin{equation} 
\mathcal{L}^g_{id}(\boldsymbol{\theta}_{\mathcal F_g})=-\frac{1}{\mathcal N}\sum_{i=1}^{\mathcal N}log(p({y_{i}|x_{i}^g})).
\end{equation}
Then the head-shoulder and joint branches identity training loss can be calculated as:
\begin{equation} 
\mathcal{L}^h_{id}(\boldsymbol{\theta}_{\mathcal F_h})=-\frac{1}{\mathcal N}\sum_{i=1}^{\mathcal N}log(p({y_{i}|x_{i}^h})),
\end{equation}
\begin{equation} 
\mathcal{L}^{joint}_{id}(\boldsymbol{\theta}_{\mathcal F_g}, \boldsymbol{\theta}_{\mathcal F_h})=-\frac{1}{\mathcal N}\sum_{i=1}^{\mathcal N}log(p({y_{i}|x_{i}^g \oplus x_{i}^h})),
\end{equation}
where $\mathcal{L}^{joint}_{id}(\boldsymbol{\theta}_{\mathcal F_g}, \boldsymbol{\theta}_{\mathcal F_h})$ optimizes the joint feature $\mathcal{V}^{joint}_i$ for supervising both global and head-shoulder branches. Specifically, the $p({y_{i}|x_{i}^g \oplus x_{i}^h})$ is denoted as:
\begin{equation} 
p({y_{i}|x_{i}^g \oplus x_{i}^h})=\frac{exp(W^T_{y_{i}}\times  (\mathcal V_{i}^g \oplus \mathcal V_{i}^h))}{\sum_{k=1}^{\mathcal{N}}exp(W^T_{k}\times  (\mathcal V_{i}^g \oplus \mathcal V_{i}^h))}.
\end{equation}

\textbf{Label Smoothing Regularization:} For the joint embedding vectors $\mathcal V^{joint}$ directly concatenated by $\mathcal V^{g}$ and $\mathcal V^{h}$, their the information distribution are generally inconsistent in the feature space. It leads to an increase in the prediction probability of wrong labels. Conventional cross-entropy loss with one-shot hard label only pay attention to how to produce a higher probability to predict the correct label, rather than reducing the probability of predicting the wrong label. In this work, we employ the label smoothing regularization (LSR) strategy for $\mathcal V^{joint}$ to alleviate this problem. Given a global image $x_i^g$ and its corresponding head-shoulder image $x_i^h$, we denote $y$ as their shared truth identity label. The re-assignment of the label distribution of each joint embedding vector is written as:
\begin{equation}
\label{eq6}
q_{i}=\left\{
\begin{aligned}
&1-\epsilon+\dfrac{\epsilon}{C}  &  (y=i), \\
&\dfrac{\epsilon}{C} &   (y\neq i),
\end{aligned}
\right.
\end{equation}
where $C$ indicates the number of all identity in the training set. $\epsilon$ is the weight parameter to balance the original ground-truth distribution $p({y_{i}|x_{i}^g \oplus x_{i}^h})$ and adaptive label smoothing distribution $q_{i}$. In this work, $\epsilon$ is set to 0.1. Then, the cross-entropy loss in Eq. (17) can be re-defined as,
\begin{equation} 
\mathcal{ L}^{joint}_{lsr}(\boldsymbol{\theta}_{\mathcal F_g}, \boldsymbol{\theta}_{\mathcal F_h})=-\frac{1}{\mathcal N}\sum_{i=1}^{\mathcal N}q_{i}log(p({y_{i}|x_{i}^g \oplus x_{i}^h})).
\end{equation}

By summing the identity loss and label smoothing regularization term mentioned above, we come up with the final hybrid loss function for supervising the joint branch:
\begin{equation} 
\mathcal{\tilde L}^{joint}_{id}(\boldsymbol{\theta}_{\mathcal F_g}, \boldsymbol{\theta}_{\mathcal F_h}) = \mathcal{ L}^{joint}_{id} + \omega \mathcal{ L}^{joint}_{lsr},
\end{equation}
where $\omega$ is a weight coefficient to control the contirbution of label smoothing regularization term.

\subsection{Multi-Branch Synchronous Learning Module.}
\subsubsection{Multi-Branch Synchronous Learning}
We perform multi-branch synchronous learning on person identity classes from global and head-shoulder specific branches. For one global image $x^g$ and its corresponding head-shoulder image $x^h$, we first feed them into the multi-granularity feature extraction module to obtain the highest convolutional feature maps $\mathcal V^g$ ($2048\times n$), $\mathcal V^h$ ($2048\times n$) repectively, where $n$ means the mini-batch size. Then, we perform the feature fusion by an operation of concatenation, that is, the dimension of the joint feature $\mathcal V^{joint}$ is $4096\times n$. Notice that different feature embedding vectors $\mathcal V^g$, $\mathcal V^h$ and $\mathcal V^{joint}$ have different information distributions, we employ three independent batch hard mining triplet losses $\mathcal L^{g}_t$ (Eq. (10)), $\mathcal L^{h}_t$ (Eq. (11)), $\mathcal L^{joint}_t$ (Eq. (12)) for synchronous metric learning. Besides, we also deploy an identity classification layer (i.e. \textit{synchronous learning layer}) for the joint feature to conduct synchronous classification learning. The training of each branch is supervised by the same identity class label constraint $\mathcal L_{id}^g$ (Eq. (15)), $\mathcal L_{id}^h$ (Eq. (16)) and $\mathcal L_{id}^{joint}$ (Eq. (17)) concurrently.

\subsubsection{Feature Regularisation by Synchronous Propagation}
We propose to regularize the branch-specific and therefore indirectly radiate to the entire feature learning process with multi-granularity person identity synchronization in a closed-loop. Specifically, we propagate the fused knowledge as extra feedback information to regularise the batch learning of all branch-specific branches concurrently.  Formally, as shown in Fig. 3, we utilize the fused knowledge probability $\tilde{\mathcal P}$ as the synchronous propagation signal (called as ``soft target"
) to guide the learning process of both global and head-shoulder branches, where $\tilde{\mathcal P}$ is the posterior probability of the joint feature, that is:
\begin{equation} 
\tilde{\mathcal P} = p({y_{i}|x_{i}^g \oplus x_{i}^h}).
\end{equation}

Then, we enforce an additional regularisation in Eq. (15) and Eq. (16), repectively:
\begin{equation} 
\mathcal{ L}^g_{id}(\boldsymbol{\theta}_{\mathcal F_g}, \boldsymbol{\theta}_{\mathcal F_h})=\mathcal{L}^g_{id}(\boldsymbol{\theta}_{\mathcal F_g}) + \lambda_3 \mathcal Z ( \mathcal {\tilde P},\mathcal P_g )
\end{equation}

\begin{equation} 
\mathcal{\tilde L}^h_{id}(\boldsymbol{\theta}_{\mathcal F_g}, \boldsymbol{\theta}_{\mathcal F_h})=\mathcal{L}^h_{id}(\boldsymbol{\theta}_{\mathcal F_h}) + \lambda_4 \mathcal Z ( \mathcal {\tilde P},\mathcal P_h ),
\end{equation}
where $\mathcal P_g = p({y_{i}|x_i^g})$, $\mathcal P_h = p({y_{i}|x_i^h})$. $\lambda_3$ and $\lambda_4$ are the predefined tradeoff coefficients for balancing the contributions between the two terms. $\mathcal{Z}(.)$ denotes the synchronous regularisation term which aims to calculate the Kullback-Leibler divergence between two distributions ($\mathcal {\tilde P},\mathcal P_g$), ($\mathcal {\tilde P},\mathcal P_h$):
\begin{equation} 
\mathcal{Z}(\mathcal {\tilde P},\mathcal P_g) = \frac{1}{\mathcal N}\sum_{i=1}^{\mathcal N}(\tilde p_iln(\tilde p_i)- \tilde p_iln(p_i^g)),
\end{equation}
\begin{equation} 
\mathcal{Z}(\mathcal {\tilde P},\mathcal P_h) = \frac{1}{\mathcal N}\sum_{i=1}^{\mathcal N}(\tilde p_iln(\tilde p_i)- \tilde p_iln(p_i^h)).
\end{equation}

\begin{algorithm}[t]
	\caption{\textbf{:} Multi-Granularity Feature Learning} 
	\hspace*{0.02in} {\bf Input:} 
	Input AGM global images $X^g=\{x_i^g\}_{i=1}^{\mathcal N}$; \\
	\hspace*{0.46in}Corresponding labels $Y=\{y_i\}_{i=1}^{\mathcal N}$;\\
	\hspace*{0.46in}Training iterations $\mathcal{I}$; learning rate $r$; batch size $\mathcal N$.\\
	\hspace*{0.02in} {\bf Initialisation:} Initialized network parameters $\boldsymbol{\theta}_{\mathcal F_g}'$ and $\boldsymbol{\theta}_{\mathcal F_h}'$; \\
	\hspace*{0.02in} {\bf Output:} 
	Network parameters $\boldsymbol{\theta}_{\mathcal F_g}$ and $\boldsymbol{\theta}_{\mathcal F_h}$.
	\begin{algorithmic}[1]
		\State \textbf{for} iteration i in $\mathcal I$; \\
		\hspace*{0.20in}Get head-shoulder training samples: $X^h=\{x_i^h\}_{i=1}^{\mathcal N}$;\\		
		\hspace*{0.20in}Feedforward global and head-shoulder image inputs (Eq. (\textcolor{red}{7}), Eq. (\textcolor{red}{8}) and Eq. (\textcolor{red}{9}));\\
		\hspace*{0.20in}Multi-granularity feature fusion (Eq. (\textcolor{red}{10}));\\
		\hspace*{0.20in}Update global network parameters $\boldsymbol{\theta}_{\mathcal F_g}$ :
		
		$\boldsymbol{\theta}_{\mathcal F_g} \gets \boldsymbol{\theta}_{\mathcal F_g} - r*\nabla( \mathcal{L}_{t}^g(\boldsymbol{\theta}_{\mathcal F_g}))$ \\
		\hspace*{0.20in}Update head-shoulder network parameters $\boldsymbol{\theta}_{\mathcal F_h}$:
		
		$\boldsymbol{\theta}_{\mathcal F_h} \gets \boldsymbol{\theta}_{\mathcal F_h} - r*\nabla( \mathcal{L}_{t}^g(\boldsymbol{\theta}_{\mathcal F_h}))$ \\
		\hspace*{0.20in}Update joint network parameters $\boldsymbol{\theta}_{\mathcal F_g}$ and $\boldsymbol{\theta}_{\mathcal F_h}$:
		
		$(\boldsymbol{\theta}_{\mathcal F_g},\boldsymbol{\theta}_{\mathcal F_h}) \gets (\boldsymbol{\theta}_{\mathcal F_g},\boldsymbol{\theta}_{\mathcal F_h}) - r*\nabla(\mathcal{\tilde L}^{joint}_{id}(\boldsymbol{\theta}_{\mathcal F_g}, \boldsymbol{\theta}_{\mathcal F_h})$
		$+\mathcal{\tilde L}^g_{id}(\boldsymbol{\theta}_{\mathcal F_g}, \boldsymbol{\theta}_{\mathcal F_h})+ \mathcal{\tilde L}^h_{id}(\boldsymbol{\theta}_{\mathcal F_g}, \boldsymbol{\theta}_{\mathcal F_h})+\mathcal{L}_{t}^{joint}(\boldsymbol{\theta}_{\mathcal F_g}, \boldsymbol{\theta}_{\mathcal F_h}))$ 
		\State \textbf{end}
		\State \textbf{return} Network parameters $\boldsymbol{\theta}_{\mathcal F_g}$ and $\boldsymbol{\theta}_{\mathcal F_h}$.

	\end{algorithmic}
\end{algorithm}

\textbf{Overall Loss Function:} Combining these individual losses, we finally define the total loss for the overall network as follows:
\begin{equation} 
\begin{split}
\mathcal{L}_{total} = &\mathcal{\tilde L}^g_{id}(\boldsymbol{\theta}_{\mathcal F_g}, \boldsymbol{\theta}_{\mathcal F_h})+ \mathcal{\tilde L}^h_{id}(\boldsymbol{\theta}_{\mathcal F_g}, \boldsymbol{\theta}_{\mathcal F_h})+\mathcal{\tilde L}^{joint}_{id}(\boldsymbol{\theta}_{\mathcal F_g}, \boldsymbol{\theta}_{\mathcal F_h}) \\ +&\mathcal{L}_{t}^g(\boldsymbol{\theta}_{\mathcal F_g}) + \mathcal{L}_{t}^h(\boldsymbol{\theta}_{\mathcal F_h})+ \mathcal{L}_{t}^{joint}(\boldsymbol{\theta}_{\mathcal F_g}, \boldsymbol{\theta}_{\mathcal F_h}).
\end{split}
\end{equation}
To this end, we introduce a framework for visible-infrared person Re-ID, in which $\mathcal{\tilde L}^g_{id}(\boldsymbol{\theta}_{\mathcal F_g}$, $\boldsymbol{\theta}_{\mathcal F_h})$ and $ \mathcal{\tilde L}^h_{id}(\boldsymbol{\theta}_{\mathcal F_g}, \boldsymbol{\theta}_{\mathcal F_h})$  aim to propagate the learned fused knowledge back to individual specific branches to regulate their mini-batch iterative learning behaviour together. Meanwhile, hard mining triplet loss $\mathcal{L}_{t}^g(\boldsymbol{\theta}_{\mathcal F_g})$, $\mathcal{L}_{t}^h(\boldsymbol{\theta}_{\mathcal F_h})$ and $\mathcal{L}_{t}^{joint}(\boldsymbol{\theta}_{\mathcal F_g}, \boldsymbol{\theta}_{\mathcal F_h})$ attempt to enhance the discriminability of learned features. Note that $\mathcal{\tilde L}^{joint}_{id}(\boldsymbol{\theta}_{\mathcal F_g}, \boldsymbol{\theta}_{\mathcal F_h})$ is calculated by both ``soft target" and groundtruth one-hot ``hard target" and is used to update the whole network parameter. The overall algorithm of training the proposed model
is presented in Algorithm 1.

\section{Experiments}
In this section, we present a detailed analysis and measure our method against other VI-ReID approaches on two available public datasets (SYSU-MM01 and RegDB). 
\subsection{Datasets and Evaluation Metric}

\textbf{Datasets.} \textbf{SYSU-MM01} \cite{Wu2017} is a challenging large-scale cross-modality dataset collected at Sun Yat-sen university. It contains images captured by six cameras (two near-infrared and four visible sensors), including both indoor and outdoor environments. Statistically, SYSU-MM01 dataset contains a total of 30,071 visible images and 15,792 thermal images of 491 person identities, where each identity is captured by at least two modality cameras. Follow the \cite{Wu2017}, we conduct our experiment on two different evaluation modes, \textit{i.e.}, all search and indoor-search mode. For all-search mode, 3,803 thermal images from cameras 3 and 6 are used for query, and 301 visible images are randomly selected from cameras 1, 2, 4, and 5 are formulated as gallery set. For indoor-search, only the images captured by two indoor cameras are used.

\textbf{RegDB} \cite{Nguyen2017} is collected by a pair of aligned far-infrared and visible camera systems. It is composed of 8,240 images of 412 identities, with 206 identities for training and the rest for testing. For each identity, 10 images are captured by the visible camera, and 10 images are obtained by the thermal camera. We following a previously developed evaluation protocol \cite{Liu2021} that randomly splits the dataset into two halves and alternatively uses all visible/thermal images as the gallery set.

\textbf{Evaluation Metric.} To evaluate the cross-modality Re-ID system performance, we adopt the widely used Cumulated Matching Characteristics (CMC) curve and mean Average Precision (mAP) for performance evaluation. In addition, we also introduce mean inverse negative penalty (mINP) metric in this work to measure the retrieval performance. Specifically, CMC (rank-k matching accuracy) measures the probability that a query object appears in the target lists (top-k retrieved results). mAP measures the retrieval performance via calculating average of the maximum recalls for each class in multiple types of tests. Finally, mINP evaluates the ability of Re-ID system to retrieve the hardest correct match, providing a strong
supplement for CMC and mAP.

\begin{table*}
	\setlength{\abovecaptionskip}{-0.2cm} 
	\setlength{\belowcaptionskip}{0.5cm}
	\caption{Ablation study of each component with four different types of training/testing sets on the large-scale SYSU-MM01 dataset. `RGB-IR' means the RGB to infrared modality dataset, `RGB-IR+GN' means the RGB to grayscale modality dataset, `Gray-IR' means the grayscale to infrared modality dataset and `Gray-IR+GN' means the grayscale to grayscale modality dataset. In addition, `HS' denotes using head-shoulder information to assit feature learning and `SLS' denotes the synchronous learning strategy. GeM pooling method is used in this experiment. }
	\label{tab:1}       
	\begin{center}
		\renewcommand{\arraystretch}{1.07}
		\begin{tabular}{l|ccccc|ccccc}
			\toprule
			Modes & \multicolumn{5}{c}{All Search} & \multicolumn{5}{c}{Indoor Search} \\
			\hline
			Method &Rank-1  &Rank-10 &Rank-20 &mAP &mINP &Rank-1  &Rank-10 &Rank-20 &mAP &mINP \\
			\hline
			RGB-IR (Baseline-A) &60.35 &91.19 &95.98 &56.31 &43.70 &65.81 &95.83 &99.50 &71.65 &67.32  \\
			RGB-IR+HS &\textbf{63.79} &90.93 &95.95 &61.38 &47.93  &\textbf{68.43} &95.88 &99.50 &\textbf{73.41} &\textbf{69.07} \\
			RGB-IR+HS+SLS &63.48 &\textbf{92.90} &\textbf{97.82} &\textbf{62.34} &\textbf{48.96} &67.62 &\textbf{96.11} &\textbf{99.55} &72.80 &68.24 \\
			\hline
			RGB-IR+GN &61.35 &91.24 &96.24 &59.72 &46.99 &66.08  &95.15 &99.23 &71.10 &67.73  \\
			RGB-IR+GN+HS &64.66 &93.14 &\textbf{97.69} &62.45 &49.28  &68.54 &96.42 &\textbf{99.64} &74.02 &69.65 \\
			RGB-IR+GN+HS+SLS &\textbf{65.42} &\textbf{93.64} &97.55 &\textbf{62.82} &\textbf{49.81}  &\textbf{69.44} &\textbf{96.97} &99.46 &\textbf{75.73} &\textbf{69.77} \\
			\hline
			\hline
			Gray-IR (Baseline-B)  &63.35 &\textbf{92.77} &97.13 &58.59 &44.49 &69.34 &\textbf{97.51} &\textbf{99.50} &74.51 &70.12 \\
			Gray-IR+HS  &63.24 &91.77 &\textbf{97.29} &61.27 & \textbf{49.66} &70.83 &96.69 &98.73 &75.28 &70.78 \\
			Gray-IR+HS+SLS  &\textbf{64.03} &91.98 &96.16 &\textbf{61.55} &48.99 &\textbf{73.91} &96.92 &99.18 &\textbf{77.47} &\textbf{72.87}\\
			\hline
			Gray-IR+GN (AGM)  &65.58 &95.42 &98.82 &62.12 &47.74 &69.38 &95.06 &97.46 &73.32 &68.04 \\
			Gray-IR+GN+HS  &67.13 &95.61 &98.55 &64.11 &50.49 &73.87 &96.92 &99.09 &77.25 &72.74 \\
			Gray-IR+GN+HS+SLS  &\textbf{69.63} &\textbf{96.27} &\textbf{98.82} &\textbf{66.11} &\textbf{52.24} &\textbf{74.68} &\textbf{97.51} &\textbf{99.14} &\textbf{78.30} &\textbf{74.00} \\
			\bottomrule
		\end{tabular}
	\end{center}	
\end{table*}

\subsection{Implementation Details}
The proposed method is implemented in PyTorch and trained on two 24GB NVIDIA TITAN RTX GPU for acceleration. Before the
training stage, all global input images are first resized to 288 $\times$ 144 and corresponding head-shoulder images are resized to 128$\times$144 to obtain sufficient context information from person images. Then we augment training samples with two data augmentation approaches, \textit{i.e.}, Random Cropping and Random Erasing. The total number of training epochs is 80, and the batch size is set to 64. We start training with learning rate 0.01 and linearly increase to 0.1 in the first 10 epochs, then we keep the same value setting until reaching to 20 epochs. In the following 60 epochs, learning rate is set to 0.01 for the first 30 epochs and 0.001 for another 30 epochs. We adopt the SGD optimizer with a weight decay of 5$\times$$10^{-4}$ and a momentum of $0.9$ to update the parameters of the network.  The hype-parameters $\lambda_1$ and $\lambda_2$ are set to 10 and 5, repectively. We set the margin parameter $\xi$ to 0.3 in Eq. (11), Eq. (12) and Eq. (13) for the batch hard triplet loss. The dimensions of the last classification layer are 395 for SYSU-MM01 and 206 for RegDB.

\subsection{Ablation Study}
\begin{table}[t]
	\centering
	\setlength{\abovecaptionskip}{0cm} 
	\setlength{\belowcaptionskip}{0.5cm}
	\caption{Comparison of components over baseline model (Baseline-A) using GeM pooling. Rank-1 (\%), mAP (\%) and mINP (\%) are reported.}
	\label{Tab03}
	\renewcommand{\arraystretch}{1}
	\begin{tabular}{ccccc|ccc}	
		\toprule
		\multirow{2}*{Base}	&\multirow{2}*{Gray}  &\multirow{2}*{GN}   &\multirow{2}*{HS} &\multirow{2}*{SLS} & \multicolumn{3}{c}{All search} \\	
		\cmidrule(r){6-8}	 	
		& & & & &Rank-1   &mAP &mINP \\			
		\hline	
		\checkmark &-- &-- &--&--&60.35 &56.31 &43.70 \\	
		\checkmark &\checkmark& -- &--&--&63.35 &58.59 &44.49 \\
		\checkmark &\checkmark&\checkmark &--&--&65.58 &62.12 &47.74\\
		\checkmark &\checkmark& \checkmark &\checkmark&--&67.13 &64.11 &50.49\\
		\checkmark &\checkmark &\checkmark&\checkmark& \checkmark &69.63 &66.11 &52.24 \\
		
		\bottomrule		
	\end{tabular}
	
\end{table}

In this section, we investigate the effectiveness of each component in our proposed framework by conducting a series of experiments on the challenging SYSU-MM01 dataset under both all search and indoor search modes.
\subsubsection{Effectiveness of the Aligned Grayscale Modality}
We first study the effectiveness of our proposed aligned grayscale modality strategy (denoted by `AGM' in TABLE 1 and `Base+Gray+GN' in TABLE 2). In TABLE 1, we utilize the base model w/wo AGM module (Basline-A and AGM) as the baseline for evaluating other components to see how their performance would change. All other settings between the Basline-A and AGM including the network architecture are consistent. 
Comparing results in row 3 and row 12, we can see that the rank-1, mAP and mINP accuracy of AGM go beyond the `Baseline-A' by 5.23\%, 5.81\% and 4.04\% under the all-search mode. This indicates that eliminating modality discrepancy is
critical to boost VI-ReID performance. Then, we add other modules proposed in this work on the top of `Baseline-A' and `AGM', repectively. As expected from the reported results in rows 12-14, our AGM based model also shows very competitive performance improvement against the general RGB-infrared based model (rows 3, 4 and 5), where the improvement of rank-1 accuracy from 65.58\% to 69.63\% on `AGM' versus 60.35\% to 63.48\% on `Basline-A'.

\begin{table}[t]
	\centering
	\setlength{\abovecaptionskip}{0cm} 
	\caption{Comparison (\%) to related cross-modality image-level Re-ID methods under the same training/tesing setting. Global average pooling is used. }
	\label{Tab03}
	\renewcommand{\arraystretch}{1.07}
	\begin{tabular}{l|ccccc}
		\toprule
		& \multicolumn{5}{c}{SYSU-MM01(All-search)}  \\ \hline
		{Methods}  & Rank-1 & Rank-10 & Rank-20 &  mAP  &    mINP    \\ \hline
		CoSiGAN \cite{Zhong2020}  & 35.55  & 81.54  &    90.43     & 38.33  & --   \\
		AlignGAN \cite{Wang2019}   & 42.40  &85.00  &93.70     & 40.70  &--   \\
		D$^2$RL\cite{Wang2019a} & 28.90  & 70.60 &    82.40     & 29.20  & --  \\
		\hline
		
		G-modal\cite{Ye2020a} & 47.80  & 86.56 &    94.12    & 45.99  & --    \\
		X-modal\cite{Li2020}    & 49.92  & 89.79 &    95.96     & 50.73  & --    \\
		S-modal\cite{Wei2021}  & 59.97  & -- &    --     & 56.01  & --    \\
		\hline
		
		AGM (Ours) & 62.35  & 92.77 &    97.13     & 58.59  & 44.49   \\
		\bottomrule
	\end{tabular}
	
\end{table}

In addition, note that AGM reformulates visible-infrared dual-mode learning as the gray-gray single-modality learning paradigm, that falls into the same category with image generation-based methods. Therefore, to validate the superiority of our proposed AGM, we further report comparison results with other classic image-level methods in TABLE 3. Here, we only use global branch to extract person features and supervise the model with standard softmax and triplet losses for the fairness of comparison. From the TABLE 3, we have following 
observation. (1) \textit{vs modality transfered methods (rows 3-5):} AGM obtains siginificantly competitive results, of which both rank-1 accuracy and mAP value have increased by more than 20.17\%. This indicates using GAN technique to generate modality consistent images (\textit{i.e.} RGB to infrared or infrared to RGB) does significantly harm the performance and demonstrates the necessity of preserving image structure information when performing cross-modality transformation. (2) \textit{vs modality assisted methods (rows 6-8):} Specifically, `G-modal' (row 6) means using grayscale modality to assist the visible-infrared cross-modality feature learning. Similarly, `X-modal' (row 7) and `S-modal' (row 8) denotes using x modality and syncretic modality repectively. As shown in TABLE 3, introducing different auxiliary modalities obviously improve the performance of the model, especially the syncretic modality. However, training new modality images requires extra computation cost, and original modality discrepancy still remains unsolved. In contrast, AGM integrates two heterogeneous modalities into single unified modality for feature learning, effectively alleviating the modality discrepancy and improving the retrieval performance. 

\begin{figure}[tp]
	\begin{center}
		\includegraphics[width=0.99\linewidth]{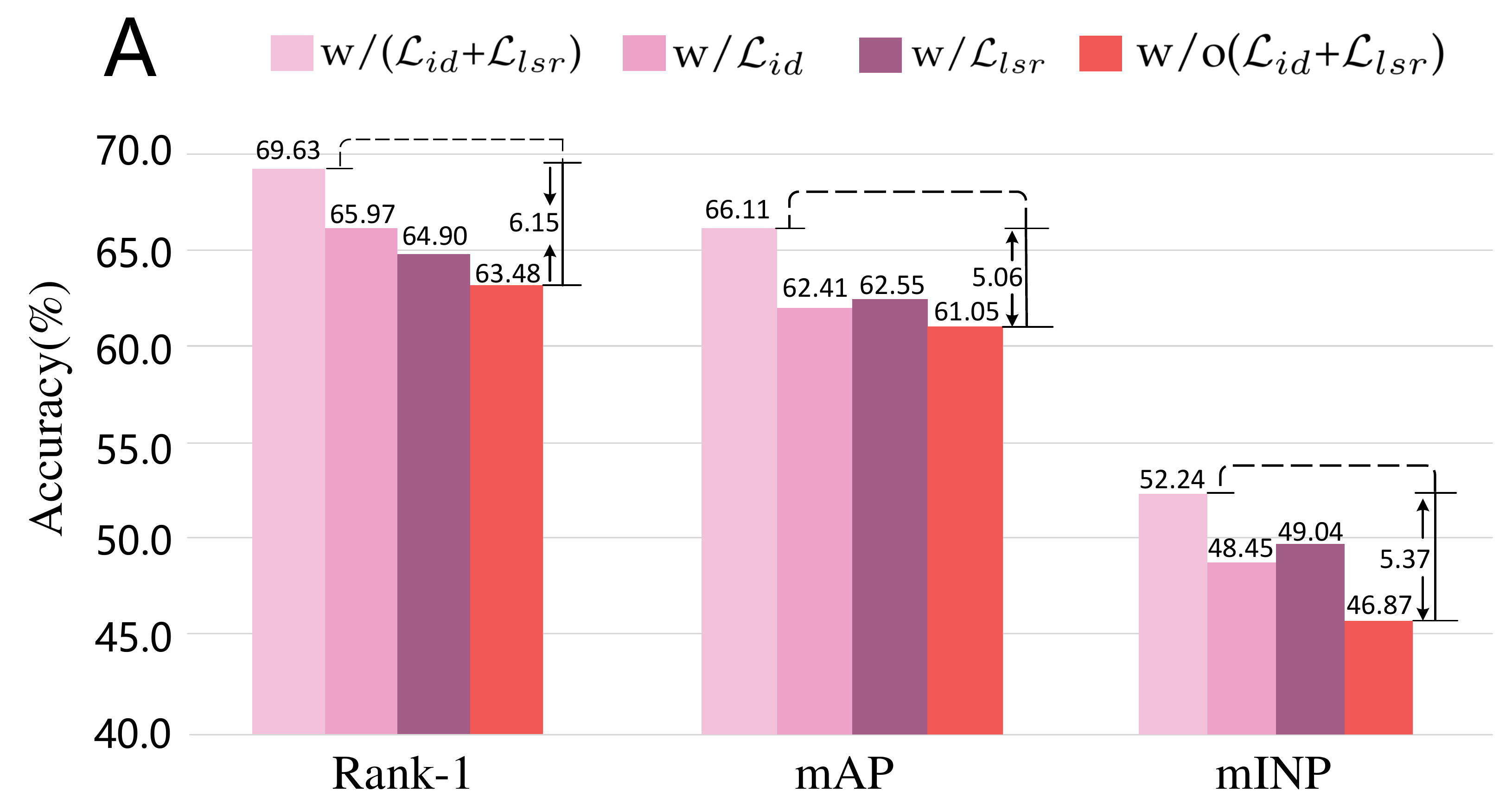}\vspace{-0.6cm}
	\end{center}
	\caption{Performance evaluation for joint branch on SYSU-MM01 dataset using different classification losses. $\mathcal{ L}_{id}$ means cross-entropy and $\mathcal{ L}_{lsr}$ denotes label smoothing regularization. \vspace{-0.4cm}}
	\label{fig:smalltarget}
\end{figure}\label{sec:introduction}

\subsubsection{Effectiveness of the Grayscale Normalization}
We evaluate how much improvement can be made by Grayscale Normalization (GN) with baseline learning objective. We first test GN under grayscale-infrared training set. From the second and third rows of TABLE 2, GN brings 2.23\% Rank-1, 3.53\% mAP and 3.25\% mINP increases in all search mode compared with the model without GN (row 2). Similar performance improvement (from 66.20\% to 69.38\%) can be observed under \textit{indoor search} mode in TABLE 1 (row 11 and 14). Then, we further test GN on the conventional RGB-infrared training set. To be fair, all settings including the network architecture are the same as grayscale-infrared training set. As shown by the results (row 3 and 6 in TABLE 1), its CMC top-1, mAP and mINP accuracy increase 1.00\%, 3.41\% and 3.29\% compared with the baseline model, which demonstrates that conducting grayscale normalization operation on infrared images helps align cross-modality feature maps to enhance the performance. Note that applying GN can significantly improve mAP and mINP metrics against CMC accuracy, this is because GN normalizes raw infrared images (with severe luminance gap) into unified grayscale images so as to improve the recognition rate of the overall classes.

\begin{figure*}[tp]
	\begin{center}
		\includegraphics[width=0.99\linewidth]{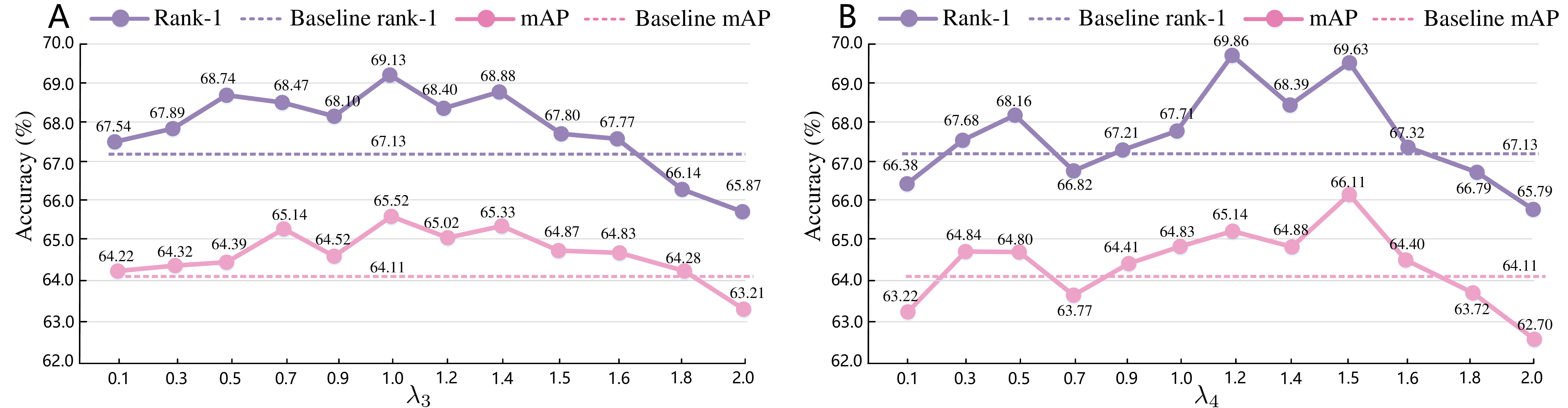}\vspace{-0.6cm}
	\end{center}
	\caption{The effect of parameter $\lambda_3$ and $\lambda_4$ on SYSU-MM01 dataset under the all-search mode. Here, $\omega$ is fixed to 1.0 for evaluation. `Baseline rank-1' and `Baseline mAP' means the rank-1 accuracy and mAP value without using synchronous learning strategy ($\lambda_3 = \lambda_4 =0.0$). \vspace{-0.4cm}}
	\label{fig:smalltarget}
\end{figure*}\label{sec:introduction}
\subsubsection{Effectiveness of the Head-shoulder Information}
This work introduces additional head-shoulder information to assist to learn discriminative feature representations, that provides a feasible research idea for future person Re-ID task. Here we conduct qualitative experiments to investigate the contribution of head-shoulder part to performance improvement.

As shown in TABLE 1, we evaluate head-shoulder module (denoted by `HS') on four different types of testing sets, \textit{i.e.}, RGB-Infrared set (RGB-IR), RGB-Gray set (RGB+GN), Gray-Infrared set (Gray-IR) and Gray-Gray set (Gray+IR). Note that when we applying head-shoulder information to assit model learning, all performance indexes (CMC curve, mAP and mINP) float with varying degrees of improvement. Especially, The rise of mAP and mINP value (3.0\% $\sim$ 5.0\%) is particularly obvious than rank-1 accuracy (- 0.1\% $\sim$ 2.0\%) on four testing sets. This demonstrates that introducing local prior knowledge \textit{(i.e.} face characteristics or head-shoulder information) is
profitable to enhance the discriminative and robust power of learned feature representations.

We also provide the comparision result when regularizing the joint branch feature using a standard one-hot cross-entropy loss, together with a label smoothing regularization. As shown in Fig. 5, using cross-entropy alone (w/$\mathcal{ L}_{id}$) improves the rank-1 accuracy from 63.48\% to 65.97\% (\textcolor{red}{$\uparrow$}2.49\%). However, replacing cross-entropy with the label smoothing regularization (w/$\mathcal{ L}_{lsr}$),
the rank-1 accuracy  decreases from 65.97\% to 64.90\% (\textcolor{red}{$\downarrow$}1.07\%). This suggests that using label smoothing regularization alone does not help much, but even decrease the
performance. When we concurrently use cross-entropy and label smoothing regularization (w/($\mathcal{ L}_{id}$+$\mathcal{ L}_{lsr}$)), the rank-1 accuracy increases sharply from 63.48\% to 69.63\% (\textcolor{red}{$\uparrow$}6.25\%).  Therefore, the fact that applying the label smoothing regularization improves over the baseline is not attributed to label smoothing alone, but to the interaction between the cross-entropy (``hard target") and label smoothing (``soft target"). By this experiment, we justify the necessity of using
label smoothing regularization to optimize the concatenate joint feature.

\begin{table}[t]
	\centering
	\setlength{\abovecaptionskip}{0cm} 
	\caption{The results of different synchronous learning losses on the SYSU-MM01 dataset. `None' means not using the synchronous learning strategy. `Specific-to-Joint' indicates the knowledge transfer from the specific branch to the joint branch. `Joint-to-Specific' indicates the knowledge transfer from the joint branch to the specific branch.}
	\label{Tab03}
	\renewcommand{\arraystretch}{1.07}
	\begin{tabular}{l|ccc|ccc}
		\toprule
		& \multicolumn{3}{c}{All search} & \multicolumn{3}{c}{Indoor search} \\ \hline
		{Settings}  & R-1 &  mAP  &     mINP     & R-1 &  mAP  &   mINP   \\ \hline
		None   & 67.13  &64.11 & 50.49    &69.75   &75.69  & 71.95  \\
		Specific-to-Joint  & 63.61  &61.11  & 47.33       &66.12   &71.21  &65.99   \\
		Joint-to-Specific  &69.63   &66.11  & 52.24        &74.68   &78.30  & 74.00    \\
		Mutual  &66.87   &63.11  & 48.86       &68.89   &72.75  &67.59     \\
		\bottomrule
	\end{tabular}
	
\end{table}

\subsubsection{Effectiveness of the Synchronous Learning Strategy}
In the synchronous learning process (Subection 3.3), the global and head-shoulder information are concatenated into the high-dimensional fusion features to calculate the person class probability.
Meanwhile, the calculated probability is utilized as the teacher signal to guide the learning process of specific branches. To evaluate the effectiveness and necessity of the synchronous learning strategy (SLS), we first design control groups under four types of testbed with or without `SLS'. As shown in TABLE 1, `SLS' brings 0.42\% $\sim$ 1.47\% performance improvement of mAP and mINP metrics, but some fluctuations using rank-1 index (-0.31\% $\sim$ 2.50\%). This indicates SLS can siginificantly improve the system's ability of retrieving all the relevant images. 
It is noteworthy that when performing SLS in the Gray-IR+GN (AGM) set (rows 13-14), all performance evaluation indexes (CMC, mAP and mINP) perform the best. This suggests that AGM is indeed beneficial for learning discriminative features.

Second, we also provide the result when using different knowledge transfer objectives to evaluate the synchronous learning process. From the reported results in TABLE 4, we can observe that only `Joint-to-Specific' setting outperforms the baseline (`None'), of which the rank-1 accuracy increases from 67.13\% to 69.63\%, the mAP value raises from 64.11\% to 66.11\% and mINP raises from 50.49\% to 52.24\%. However, using the other two settings (`Specific-to-Joint' and `mutual') do not help much, or even contribute  to severe performance degradation on full-camera systems. For instance, the rank-1 accuracy drops from 67.13\% to 63.61\% when using `Specific-to-Joint' setting for synchronous learning process. This is because the gobal and head-shoulder branches are conducted as an asynchronous learning scheme with a fame. If we treat specific branches as the target distribution, such the learned asynchronous knowledge will propagate to the joint branch, thereby destroying the performance.

\subsection{Parameter Analysis}
We analyze some important parameters of LSR and SLS introduced in Section 3.2.3 and Section 3.3.2. Once validated, the same parameters are fixed for other experiments.

\textbf{Label Smoothing Regularization Analysis.}
To evaluate the label smoothing regularization parameter $\omega$, we first fix $\lambda_3$ and $\lambda_4$ to 1.0 and adjust $\omega \in [0,1]$. The results are listed in TABLE 5.  From the table, we can observe that \textit{1) In SYSU-MM01 dataset},  both mAP and mINP performance rise with increasing of $\omega$ and achieve the highest performance at 1.0. This indicates the
``soft targets" and ``hard targets" should contribute equally to the learning process of the joint branch. 2) \textit{In RegDB dataset}, both rank-1, mAP performance first show an upward trend and achieve peak performance at 0.7. After that, mINP
performance drops drastically while rank-1 and mAP performance show
a downward trend with small fluctuations. This indicates the RegDB dataset is more sensitive to parameter $\omega$ the SYSU-MM01 dataset. From the above analysis, it is supposed to set $\omega=1.0$ for SYSU-MM01 dataset and  $\omega=0.7$ for RegDB dataset.

\textbf{Synchronous Learning Strategy Analysis.}
Two weighting parameters, $\lambda_3$ and $\lambda_4$, are involved in our synchronous learning module. Note that we fix one parameter value and change the other parameter in a value range for evaluation. Specifically, when evluating the parameter $\lambda_3$, we first assign a fixed value to $\lambda_4$ and then adjust  $\lambda_3 \in [0,2]$ to observe performance changes. Experimental results on SYSU-MM01 dataset are presented in Fig. 6.  From these results, we have several observations as follows.

First, different weighting parameters contirbute to different effects on model training. In Fig. 6 (a) and Fig. 6 (b), as the parameter value changes, the performance curve of $\lambda_4$ fluctuates drastically, while the performance curve of $\lambda_3$ remains relatively stable.
This demonstates that our model is more sensitive to $\lambda_4$ than $\lambda_3$. Therefore, how to balance the contribution of the introduced head-shoulder information is very important for model synchronous learning.

Second, our model favors a relatively small value for $\lambda_3$ and a large value for $\lambda_4$. From Fig. 6 (a), we can find that both rank-1 and mAP performance upgrade with increasing of $\lambda_3$ and achieve peak values
at 1.0. After that, they show a downward trend. And, in Fig. 6 (b), $\lambda_4$ varies in a similar way with $\lambda_3$, but occuring some fluctuation in value range [0.5, 0.9]. In addition, $\lambda_4$ meets two peak point at 1.2 and 1.5, repectively. Though when $\lambda_4 =1.2$, the rank-1 accuracy is the highest, its mAP value is lower than $\lambda_4 =1.5$ (65.14\% versus 66.11\%). mAP provides a comprehensive assessment of a system's ability, thus we choose the larger value: $\lambda_4=1.5$ for the experiment.
Based on the above analysis, the final value of weighting parameters are setted as: $\lambda_3=1.0$ and $\lambda_4=1.5$.

\begin{table}[t]
	\centering
	\setlength{\abovecaptionskip}{0cm} 
	\caption{ The effect of parameter $\omega$ on SYSU-MM01 and RegDB datasets. $\lambda_3$ and $\lambda_4$ are initialized to 1.0 in this experiment. Note that $\omega$ is used to balance the contributions between hard target cross-entropy loss and soft target label smoothing loss. Rank-1, mAP and mINP (\%) are reported.}
	\label{Tab03}
	\renewcommand{\arraystretch}{1.07}
	\begin{tabular}{c|c|ccc|ccc}
		\toprule
		&&\multicolumn{3}{c}{SYSU-MM01} &\multicolumn{3}{c}{RegDB}  \\ \hline
		Loss &$\omega$  & R-1 &  mAP  &     mINP   & R-1 &  mAP  &     mINP     \\ \hline
		\multirow{6}*{$\mathcal{\tilde L}^{joint}_{id}$} 
		&0.1   & 65.90  &64.46  &50.01  &  85.44  &80.26 & 69.29    \\
		&0.3  & 66.37  &63.38 & 49.30   &85.87  &80.80 & 69.04      \\
		&0.5 &\textbf{68.96}   &64.97  &50.71   & 86.60  &79.92 & 67.54        \\
		&0.7  & 67.26  &64.52  &51.09   & \textbf{87.09}  &\textbf{81.24} & \textbf{69.76}         \\
		&0.9  & 68.68  &65.87  &51.77   & 85.34  &78.14 & 63.68         \\
		&1.0  & 68.75  & \textbf{66.01} & \textbf{52.01}  & 85.78  &79.05 & 65.27         \\

		\bottomrule
	\end{tabular}
	
\end{table}

\begin{table*}[t]
	\setlength{\abovecaptionskip}{-0.2cm}
	\caption{Comparison with the state-of-the-art methods under all-search and indoor-search modes on SYSU-MM01 dataset.}
	\label{tab:1}
	\begin{center}
		
		\renewcommand{\arraystretch}{1.10}
		\begin{tabular}{l|c|ccccc||cccccc}	
			\toprule
			&	 & \multicolumn{5}{c}{All-search} & \multicolumn{5}{c}{Indoor-Search} \\
			
			\hline
			
			Method & Venue	& Rank-1     &  Rank-10   &   Rank-20   &  mAP &  mINP
			
			& Rank-1      &  Rank-10   &   Rank-20   &  mAP &  mINP
			\\
			\hline
			
			Two-Stream \cite{Wu2017}&ICCV2017  & 11.65&  47.99 &  65.50  & 12.85 &- & 15.60&  61.18  & 81.02  & 21.49  &-\\
			One-Stream \cite{Wu2017}&ICCV2017 & 12.04&  49.68  & 66.74  & 13.67 &- &16.94&63.55&82.10&22.95 &-     \\
			Zero-Pad \cite{Wu2017}  &ICCV2017 &14.80&54.12&71.33&15.95 &- & 20.58& 68.38  & 85.79  & 26.92 &-     \\
			cmGAN \cite{Dai2018} &IJCAI2018 & 26.97&  67.51  &  80.56  & 31.49 &-   & 31.63& 77.23  & 89.18  & 42.19 &-  \\
			eDBTR \cite{Ye2019} &TIFS2020 & 27.82&  67.34  &  81.34  & 28.42  &-   & 32.46& 77.42  & 89.62  & 42.46 &-  \\	
			D$^{2}$RL \cite{Wang2019a} &CVPR2019 & 28.90&  70.60  & 82.40  & 29.20 &-   & -&  -  &  -  & - &-  \\
			CoSiGAN \cite{Zhong2020} &ICMR2020 & 35.55& 81.54 & 90.43 & 38.33 &-   &- &- & - &- & - \\
			MSR \cite{Feng2019}  &TIP2020 & 37.35& 83.40 & 93.34  & 38.11 &-   & 39.64&  89.29 & 97.66 & 50.88 &-  \\
			AlignGAN \cite{Wang2019} &ICCV2019 & 42.40&  85.00  &  93.70  & 40.70 &-   & 45.90&  87.60  & 94.40  & 54.30 &-  \\
			X-Modal \cite{Li2020} &AAAI2020 & 49.92 & 89.79 & 95.96 & 50.73 & -  &- &- & - &- &-  \\
			FBP-AL \cite{Wei2021a} &TNNLS2021 & 54.14&86.04 & 93.03 & 50.20 &-   &- &- & - &- & - \\
			LLM \cite{Feng2021} &ECCV2020 & 55.25 & 86.09 & 92.69 & 52.96 &- &59.65 &90.85 &95.02 &65.46 &-\\
			NFS \cite{Chen2021} &CVPR2021 & 56.91 &91.34 &96.52 &55.45 &-  &62.79 &96.53 &99.07 &69.79   &-  \\
			VSD\cite{Tian2021} &CVPR2021 &60.02 &94.18 &98.14 &58.80 &- &66.05 &96.59 &99.38 &72.98 &-\\
			cm-SSFT \cite{Lu2020} &CVPR2020 & 61.60 &89.20 & 93.90 &63.20 &- &70.50 &94.90 &97.70 &72.60 &-\\
			GLMC\cite{Zhang2021a} &TNNLS2021 & 64.37&93.90 & 97.53  &63.43 &- &67.35& 98.10  &\textbf{99.77}  & 74.02 &-\\
			MC-AWL\cite{Ling2021} &IJCAI2021 & 64.82&- & -  &60.81 &- &-& -  &-  &- &-\\
			SMCL \cite{Wei2021} &ICCV2021 & 67.39 &92.87 & 96.76 &61.78 &-   &68.84 &96.55 &98.77 &75.56 &-  \\
			\hline
			AGW \cite{AYe2020} &TPAMI2021 & 47.50 &84.39 & 92.14 &47.65 &35.30   &54.17 &91.14 &95.98 &62.97 &59.20  \\
			DDAG \cite{Ye2020b} &ECCV2020 & 54.75 & 90.36 & 95.81 & 53.02 &39.62 &61.02 &94.06 &98.41 &67.98 &62.61\\
			IMT \cite{Xia2021} &Neuro2021 & 56.52 & 90.26 & 95.59 & 57.47 &38.75  &68.72 &94.61 &97.42 &75.22  &64.22 \\
			HTL \cite{Liu2020} &TMM2020 & 61.68 &93.10 & 97.17 &57.51 &39.54   &63.41 &91.69 &95.28 &68.17 &64.26  \\
			MCLNet \cite{Hao2021} &ICCV2021 &65.40 &93.33 &97.14 &61.98 &47.39 &72.56 &96.98 &99.20 &76.58 &72.10\\
			\hline
			\hline
			AGMNet (Ours) &This work  &\textbf{69.63} &\textbf{96.27} &\textbf{98.82} &\textbf{66.11} &\textbf{52.24} &\textbf{74.68} &\textbf{97.51} &99.14 &\textbf{78.30} &\textbf{74.00} \\
			
			\bottomrule
			
		\end{tabular}
	\end{center}

\end{table*}

\subsection{Comparison to the State-of-the-Art}
We compare the performance of the proposed AGM with state-of-the-art methods on two cross-modality benchmark datasets: SYSU-MM01 \cite{Nguyen2017}  and RegDB \cite{Wu2017}. We use a single query, and do not use any post-processing techniques (e.g., re-ranking).

\subsubsection{Performance Comparisons on SYSU-MM01}
According to the properties of the solutions, the comparison methods can be divided into two groups:  GAN-based (\textit{i.e.} cmGAN \cite{Dai2018}, D$^{2}$RL \cite{Wang2019a}, CoSiGAN \cite{Zhong2020}, AlignGAN \cite{Wang2019}, X-Modal \cite{Li2020}, \textit{etc}.) and shared feature learning (\textit{i.e.} eDBTR \cite{Ye2019}, AGW \cite{AYe2020}, FBP-AL \cite{Wei2021a}, NFS \cite{Chen2021}, VSD\cite{Tian2021}, MCLNet \cite{Hao2021}, \textit{etc}.) approaches. It is worth noting that we choose more
than ten competing methods publsihed in recent two years (2020 or 2021) for comparison. This can fully prove the superiority and advanced nature of our proposed method.

Comparison results are reported in TABLE 6. We can see that our AGMNet sets a new state of the art on SYSU-MM01, achieving 69.63\% Rank- 1 accuracy, 66.11\% mAP and 52.24\% mINP under all-search mode and 74.68\% Rank-1 accuracy, 78.30\% mAP and 74.00\% mINP under indoor-search mode. Although some methods (FBP-AL\cite{Wei2021a}, GLMC\cite{Zhang2021a} and HTL \cite{Liu2020}) introduce part-based convolutional features to improve retrieval performance, AGMNet still shows meaningful performance gain in terms of Rank-1/mAP/mINP (69.63\% vs 64.37\%, 66.11\% vs 63.43\% and 52.24\% vs 39.54\%). 

SMCL \cite{Wei2021} is most similar to ours in that we both draw support from another modality to bridge the cross-modality gap. It generates the syncretic
modality with a light-weight network and learns modality-invariant representations with the triple modality interaction learning strategy. Our model on the other hand generates the aligned grayscale modality with image graying and CycleGAN \cite{Zhu2017}. It simultaneously addresses the modality dicrepancy and luminance gap problems by translating two heterogeneous modalities into one homogeneous modality. The results indicate the single modality feature is much more robust than triple modality-shared feature, \textit{i.e.} Rank-1 accuracy 69.63\% vs 67.39\% and mAP value 66.11\% vs 61.78\%.

\subsubsection{Performance Comparisons on RegDB}
We also compare in TABLE 7 our models with the state of the art methods on RegDB \cite{Nguyen2017}. Similar to the case with the results obtained on SYSU-MM01, our approach consistently outperforms current SOTAs under both evaluation
modes.  Specifically, for visible-to-thermal mode, AGMNet achieves rank-1 accuracy of 88.40\%, mAP value of 81.45\% and mINP of 68.51\%. Noting that the current top-performing method is CAJL \cite{Ye2021} published in ICCV2021, our approach distinctly improves the Rank-1 accuracy of 3.37\% (from 85.03\% to 88.40\%), mAP of 2.31\% (from 79.14\% to 81.45\%) and mINP of 3.28\% (from 65.33\% to 68.51\%). Similar improvement can be observed under thermal-to-visible mode. For instance, AGMNet beats the
SFANet \cite{Liu2021} that adopts the same backbone model and training environment by 15.20\% in terms of Rank-1 accuracy and 17.42\% in terms of mAP. Moreover, It also outperforms the best SOTA method CAJL \cite{Ye2021} by 0.59\%, 3.37\% and 4.20\% respectively in terms of Rank-1 accuracy, mAP and mINP. 

The above comparison results are consistent with those obtained on the SYSU-MM01 database. These experimental
results demonstrate the outstanding performance of AGMNet in bebefits of its ability in discovering the discriminative features for visible-infrared person Re-ID.

\begin{figure*}[tp]
	\begin{center}
		\includegraphics[width=0.99\linewidth]{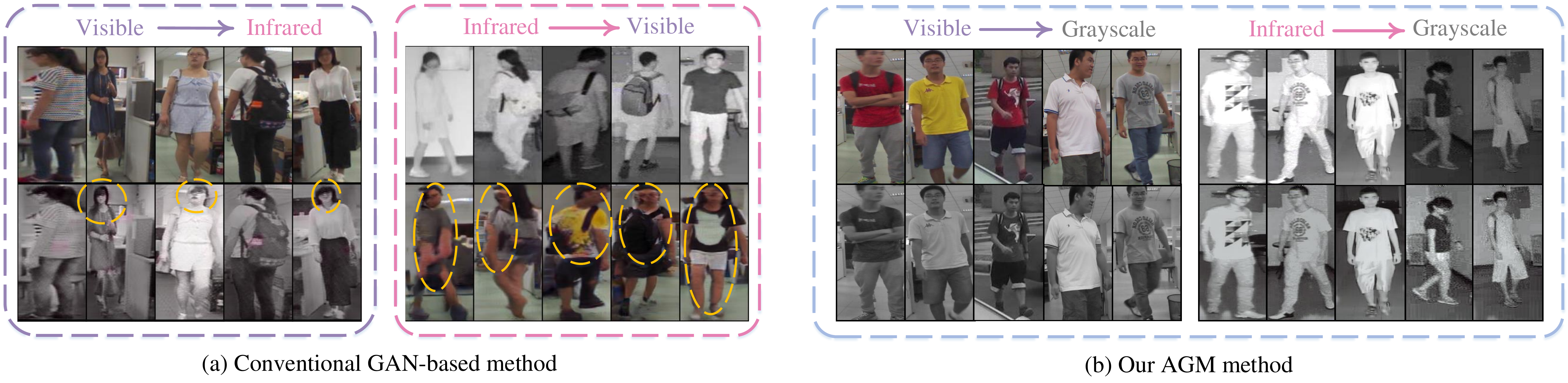}\vspace{-0.6cm}
	\end{center}
	\caption{Examples of translated images generated by vanilla image translation models such as CycleGAN (a) and our AGMNet (b).  By minimizing heterogeneous modality distances in a unified middle image space, AGM (b) significantly reduce the cross-modality
		gap. While conventional GAN-based method (a) fails to deal with this issue due to identity inconsistency during the complicated adversarial training process. \vspace{-0.4cm}}
	\label{fig:smalltarget}
\end{figure*}\label{sec:introduction}
\begin{table}
	\setlength{\abovecaptionskip}{-0.2cm}
	\caption{Comparison with the state-of-the-art methods on the RegDB datasets of different query settings. Here, only `HS' and `SLS' modules are used.}
	\label{tab:1}       
	\begin{center}
		\renewcommand{\arraystretch}{1.10}
		\begin{tabular}{l|ccccc}
			\toprule
			Setting & \multicolumn{5}{c}{Visible-Thermal} \\
			\hline
			Method&Rank-1 &Rank-10 &Rank-20 &mAP &mINP\\
			\hline
			Zero-Pad \cite{Wu2017}   &17.75&34.21&44.35&18.90 &-     \\
			eDBTR \cite{Ye2019}  & 34.62& 58.96  & 68.72  &33.46  &-     \\
			D$^{2}$RL \cite{Wang2019a}  & 43.40&  66.10  & 76.30 & 44.10 &-     \\
			CoSiGAN \cite{Zhong2020}  &47.18 &65.97 &75.29 &46.16 &-    \\
			MSR \cite{Feng2019}   &48.43 &70.32 &79.95 &48.67 &-      \\
			FBP-AL \cite{Wei2021a}  & 73.98&89.71 & 93.69 & 68.24 &-    \\
			NFS \cite{Chen2021}  &80.54 &91.96 &95.07 &72.10  &-    \\
			MPANet \cite{Wu2021}  &82.80 &- &- &80.70  &-  \\
			SMCL \cite{Wei2021} &83.93 &- &- &79.83  &-  \\
			\hline
			AGW \cite{AYe2020}  &70.05 &87.28 &92.04 &66.37 &50.19    \\
			DDAG \cite{Ye2020b}  &69.34 &86.19 &91.49 &63.46 &49.24   \\
			IMT \cite{Xia2021}  &75.49 &87.48 &92.09 &69.64 &56.30  \\
			SFANet \cite{Liu2021}  &76.31 &91.02 &94.27 &68.00  &55.92    \\
			MCLNet \cite{Hao2021} &80.31 &92.70 &96.03 &73.07 &57.39  \\
			CAJL \cite{Ye2021} &85.03 &\textbf{95.49} &\textbf{97.54} &79.14 &65.33   \\
			\hline
			AGMNet (Ours)  & \textbf{88.40}&95.10  & 96.94  &\textbf{81.45} &\textbf{68.51} \\
			\hline
			\hline
			Setting & \multicolumn{5}{c}{Thermal-Visible} \\
			\hline
			Zero-Pad \cite{Wu2017} &16.63 &34.68 &44.25 &17.82 &-\\
			eBDTR \cite{Ye2019}  &34.21 &58.74 &68.64 &32.49  &-\\
			D$^{2}$RL \cite{Wang2019a}  & 43.40&  66.10  & 76.30 & 44.10 &-     \\
			FBP-AL \cite{Wei2021a}  &70.05   &89.22 &93.88 &66.61 & -    \\
			NFS \cite{Chen2021}  &77.95 &90.45 &93.62 &69.79   &-   \\
			SMCL \cite{Wei2021} &83.05 &- &- &78.57  &-  \\
			MPANet \cite{Wu2021}  &83.70 &- &- &80.90  &-  \\
			\hline
			AGW \cite{AYe2020}  &68.83 &83.69 &88.35 &64.45 &48.74    \\
			DDAG \cite{Ye2020b}  &68.06 &85.15 &90.31 &61.80 &48.62\\
			SFANet \cite{Liu2021}  &70.15 &85.24 &89.27 &63.77   &51.97   \\
			IMT \cite{Xia2021}  &71.33 &84.52 &88.11 &66.77 &52.28 \\
			MCLNet \cite{Hao2021} &75.93 &90.93 &94.59 &69.49 &52.63  \\
			CAJL \cite{Ye2021} &84.75 &\textbf{95.33} &\textbf{97.51} &77.82 &61.56  \\
			\hline
			AGMNet (Ours) &\textbf{85.34}&94.56   &97.48  & \textbf{81.19} &\textbf{65.76} \\
			\bottomrule
		\end{tabular}
	\end{center}
	
\end{table}

\begin{table}[t]
	\centering
	\setlength{\abovecaptionskip}{0cm} 
	\caption{Quantitative results of AGM using different cross-modality baselines (\textit{i.e.} IDE \cite{Zheng2016}, PCB \cite{Sun2019} and AGW \cite{AYe2020}). Gobal average pooling method is used in this experiment. We report Rank- 1
		accuracy(\%), mAP(\%) and mINP(\%) on SYSU-MM01.}
	\label{Tab03}
	\renewcommand{\arraystretch}{1.07}
	\begin{tabular}{l|ccc|ccc}
		\toprule
		& \multicolumn{3}{c}{All search} & \multicolumn{3}{c}{Indoor search} \\ \hline
		{Methods}  & R-1 &  mAP  &     mINP     & R-1 &  mAP  &   mINP   \\ \hline
		IDE\cite{Zheng2016}    & 57.85  &53.42 & 41.20    &64.32   &69.89 & 65.01  \\
		PCB \cite{Sun2019}  &61.66   &57.84  & 42.23        &66.17   &70.48  & 66.04    \\
		AGW\cite{AYe2020}  & 60.04  & 58.84 &  46.16      &65.67   &70.91  &  66.32   \\
		\hline
		\hline
		IDE+AGM  &62.35   &58.59  &44.49     & 68.80  &73.84 &  69.16   \\
		PCB+AGM  & 65.97  &59.75  & 42.73       &71.11   & 73.67 & 67.98    \\
		AGW+AGM & 64.45 &61.26 &46.42  &69.38   &73.32  & 68.04   \\
		\bottomrule
	\end{tabular}
	
\end{table}

\subsection{Further Study and Depth Analysis}
\subsubsection{Evaluation of AGM on Different Baselines}
In fact, the proposed Aligned Grayscale Modality (AGM) can be seen as an independent data preprocessing module for cross-modality person re-identification task. It redefines visible-infrared dual-mode learning as a gray-gray single-mode learning problem. Therefore, to evaluate its effectiveness and applicability, we further test it on three commonly used baselines (\textit{i.e.} IDE\cite{Zheng2016}, PCB\cite{Sun2019} and AGW\cite{AYe2020}). For the fairness of comparison, we keep six experimental control group settings consistent during evaluation. 

The test performance on three different baselines is summarized in TABLE 8. Comparing Baseline without applying AGM, we can observe that the scores of three baselines all hover around 60\%. Interestingly, PCB achieves the highest performance on rank-1 accuracy but fails to keep its advantage in terms of mAP and mINP metrics compared to AGW. Notably, when applying AGM, all metrics on three baselines achieve a remarkable improvement. For instance, under all-search mode, IDE+AGM outperforms IDE with 5.50\% rank-1 accuracy, 5.17\% mAP and 3.29\% mINP value. And when it comes to stronger baselines PCB and AGW, PCB+AGM and AGW+AGM continuously
boosts the retrieval performance, indicating that AGM is complementary to various baselines. 
This result also shows the potential
of AGM as an independent data preprocessing method to be combined
with other baseline models.

\subsubsection{Depth Analysis of AGM and GAN-based Methods}
The Aligned Grayscale Modality (AGM) explicitly synthesizes style-consistent grayscale images in the pixel space for highly efficient modality and luminance gap elimination. The major defining difference  from other GAN-based methods (\cite{Zhong2020}, \cite{Wang2019}, \cite{Wang2019a}, \cite{Wang2020},\cite{Zhong2021}) is that we propose to utilize a unified middle modality space to reduce modality discrepancy, instead of directly generating its opposite modality. A illustration is shown in Fig. 7. This stratgey enjoys following several merits: \textbf{(1) Realistic synthetic effect.} For existing GAN-based methods, they are non-trivial to accurately choose
the suitable target for style transfer due to the separable feature statistics between visible and infrared domains. Here, AGM relys on grayscale images to conduct modality translation. Since the implicit probability distribution of infrared modality is very similar to the target probability distribution of the grayscale domain, the generator thereby can easily transfer infrared images into the target grayscale images with a high fidelity effect. 

\textbf{(2) Complete elimination of modality gap.} As shown in Fig. 7, although conventional GAN-based methods may relieve modality discrepancy to a certain extent, it incurs a considerable level of structured noise, highlighted by yellow circle. 
If these low-quality synthetic images are directly used to train an Re-ID model, a novel gap between the original data and the synthetic data will be introduced to the learning process. In contrast, AGM explores a middle modality distribution between visible and infrared domains. That is, we eliminate modality gap by simultaneously aligning the two modality distributions to the grayscale distribution. In this process, two heterogeneous modality information is integrated to one homogeneous modality information, therefore siginificantly smoothing the modality discrepancy.

\begin{table}[t]
	\centering
	\setlength{\abovecaptionskip}{0cm} 
	\caption{Quantitative results of AGM on near infrared and thermal infrared datasets. Rank- 1
		accuracy(\%), mAP(\%) and mINP(\%) are reported.}
	\label{Tab03}
	\renewcommand{\arraystretch}{1.10}
	\begin{tabular}{l|ccc|ccc}
		\toprule
		& \multicolumn{3}{c}{SYSU-MM01} & \multicolumn{3}{c}{RegDB} \\ \hline
		{Methods}  & R-1 &  mAP  &     mINP     & R-1 &  mAP  &   mINP   \\ \hline
		Baseline    & 57.85  &53.42 & 41.20    &\textbf{80.49}  &\textbf{75.68} & \textbf{61.22}  \\
		Bseline+AGM &\textbf{62.35}   &\textbf{58.59}  & \textbf{44.49}    &78.97  &73.10  & 60.15    \\
		
		\bottomrule
	\end{tabular}
	
\end{table}

\begin{figure}[tp]
	\begin{center}
		\includegraphics[width=0.99\linewidth]{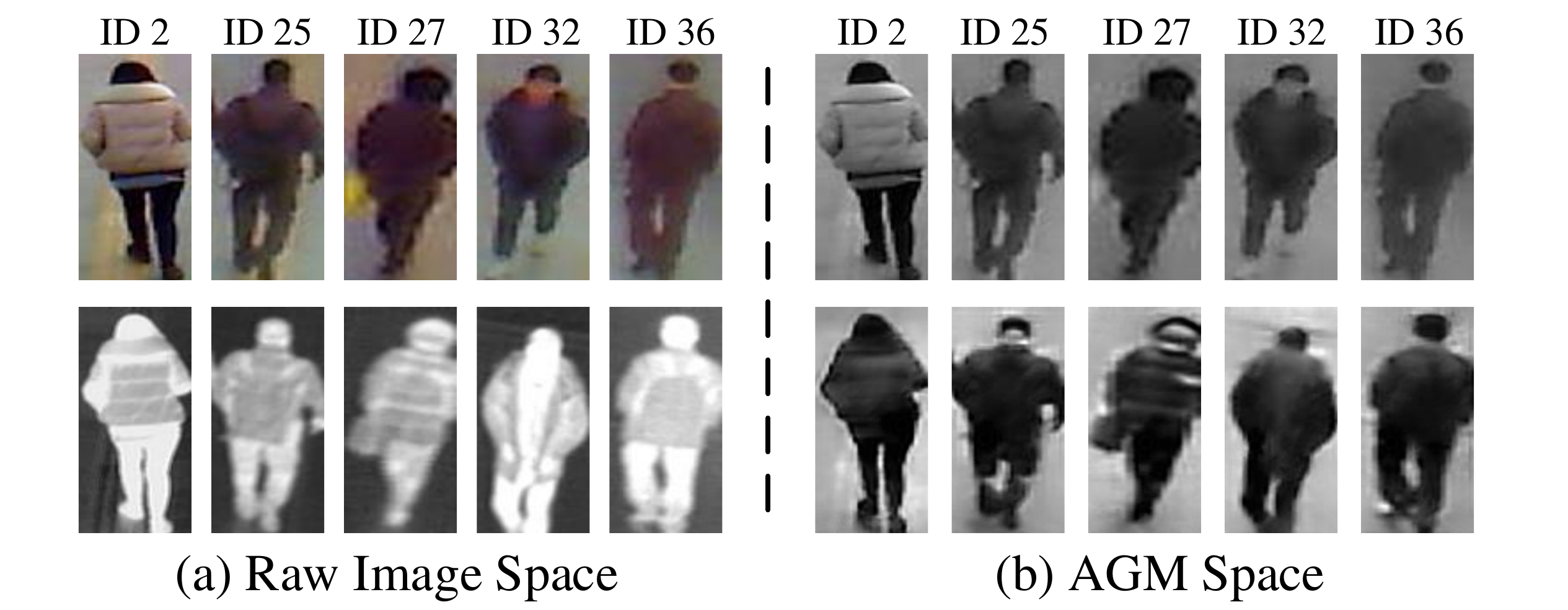}\vspace{-0.6cm}
	\end{center}
	\caption{ Contrast visualization between the raw modality images (a) and the aligned grayscale modality images (b) on RegDB dataset. Compared to raw images, AGM significantly reduces the modality discrepancy, but local feature information is thereby smoothed. \vspace{-0.4cm}}
	\label{fig:smalltarget}
\end{figure}\label{sec:introduction}

\subsubsection{Applicable Scenario of AGM}
Compared to other modality discrepancy elimination algorithms, AGM shows its unparalleled superiority on image detail preservation. 
However, some tests prove that it is not applicable to all infrared datasets. 
That is, AGM is easier to achieve a better performance on near infrared image dataset than thermal infrared dataset.
As shown in TABLE 9, using AGM alone
on RegDB (thermal infrared dataset) does not help much, or even decrease the performance from 80.49\% to 75.97\% in terms of Rank-1 accuracy. On the contrary, the performance on SYSU-MM01 (near infrared dataset) achieve significantly improvement (from 57.85\% to 62.35\%). This is because near infrared images share similar style with grayscale image, while the style discrepancy between the thermal infrared and grayscale modalities still exists.
Besides, it is worth noting that in a thermal image, person area is presented with white pixel points and other irrelevant backgrounds are presented with black pixel points. This imaging process, in fact, is equal to apply a predefined attention pattern map of itself. If we transfer thermal infrared images to grayscale style, as shown in Fig. 8, such the role of attention would be weakened and the loss outweighs the gain. 

The above analysis shows that AGM, in practice, is more suitable to near infrared datesets, such as SYSU-MM01\cite{Wu2017}, CASIA\cite{Li2007} and CASIA NIR-VIS 2.0 \cite{Li2013},\textit{ etc}. Fortunately, most of commercial infrared cameras are imaged in the form of near-infrared light, which means the poposed AGM has a very far-reaching practical application value.

\subsubsection{Head-shoulder Information vs PCB}
The proposed head-shoulder information module shares some spirit of Part-based Convolutional Baseline (PCB) by learning discriminative part-informed features. 
This is because, the head-shoulder information can be considered as a kind of part-level descriptor via proactively cropping from the global image. However, it differs significantly from PCB in the following perspectives: \textit{1) Generation way}: PCB  takes a whole image as the input and outputs a convolutional descriptor that contains part-level features. That is, it generates part-level features by partitioning the convolutional tensor. In contrast, head-shoulder information is directly generated from the original image.  The rational behind is that the head and shoulder positions contain the most discriminative fine-grained information to depict a person. \textit{2) Learning way}: PCB  calculates the cross-entropy loss for every part-level column
vector, and minimize the sum losses to optimize the network parameters. It improves the retrieval performance benifiting from its spatial alignment. In contrast, head-shoulder information aims to assit the global feature to form a stronger feature descriptor. In other words, it improves the retrieval performance due to integrate more feature map information.

\begin{figure}[tp]
	\begin{center}
		\includegraphics[width=0.99\linewidth]{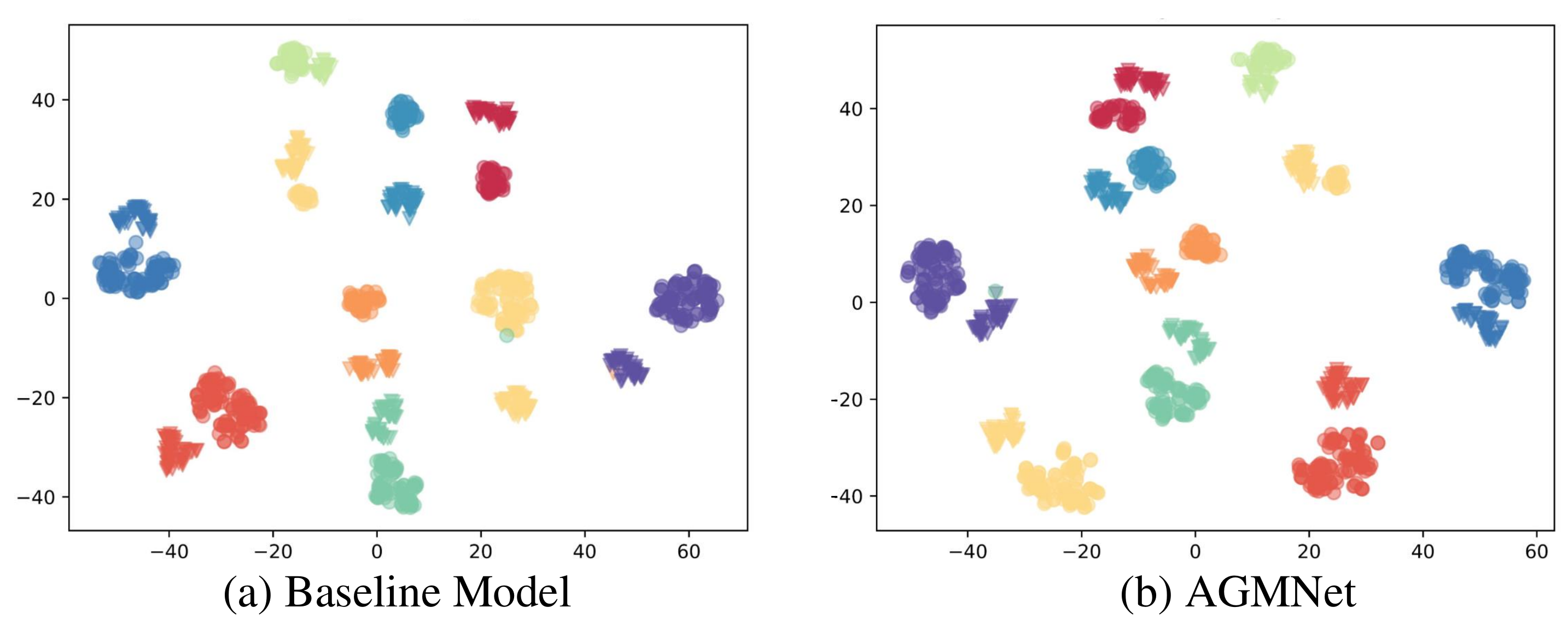}\vspace{-0.6cm}
	\end{center}
	\caption{ t-SNE visualization of the distribution of learned representations on SYSU-MM01 dataset. Each color represents an identity in the testing set. The triangles and circles represent different features extracted from the visible and infrared modalities, respectively.  \vspace{-0.4cm}}
	\label{fig:smalltarget}
\end{figure}\label{sec:introduction}

\subsubsection{Synchronous Learning vs Knowledge Distillation}
For synchronous Learning, the consensus feedback propagation can be considered as a kind of knowledge transfer via aligning relative-entropy soft targets. 
Seemingly, it may share some essence
with Knowledge Distillation (KD) that transfers between a static pre-defined teacher and a student in model distillation. However, synchronous Learning differs significantly from KD: \textit{1) Different objectives:} The distil-lation based approaches start with a powerful deep teacher network, and then train a smaller student network to mimic the teacher. The motivation behind it is how to exploit few parameters to train a model that  has the same representation capacity as the large network. On the contrary, synchronous learning stratgey aims to obtain more discriminative person representations via  multi-branch feature information interaction. \textit{2) Dynamics:} For KD, the teacher model is always a powerful pre-trained network. That is, the teacher's class probabilities are fixed during distillation. Instead, the synchronous learning strategy exploits the per-batch outputs of all student models to generate the teacher signals. Hence, it conveys additional information dynamically in an interactive manner rather than statically as KD.

\subsubsection{Visualization analysis}
Finally, we give a microscopic interpretation of AGMNet from the perspective of visualization analysis. We examine the internal features captured by baseline model and AGMNet using t-SNE, respectively.

As shown in Fig. 9(a), with baseline model, the extracted test features have significant modality discrepancy, in which feature distributions from visible and infrared modalities are fairly farther and less discriminable. Furthermore, the intra-identity modality discrepancy remains still obvious (orange and green triangles). Specifically, the distance between orange and green triangles are closer
than that between organe triangles and organe circles, which contributes to the model misjudging orange and green triangles into the same person. In contrast, as shown in Fig. 9(b), feature distributions from visible and infrared modalities are fairly closer and therefore more discriminable. This indicates that AGMNet effectively minimizes the
cross-modality gap by aligning distributions of the two modalities, where the learned features of different modalities are grouped by identity instread of modality.

\section{Conclusion}
This paper presents a novel insight of modality discrepancy elimination for visible-infrared person Re-ID task. The proposed Aligned Grayscale Modality (AGM) explicitly sets up a unified middle image space to integrate multi-modality information, that reformulates heterogeneous modality learning into homogeneous grayscale modality learning problem. Moreover, to reduce the intra-class discrepancy, we propose to utilize the head-shoulder 
information to assist global features for feature learning. 
In contrast to 
models that only employ global appearance features, the proposed AGMNet significantly learns the consensus on identity classes between global and head-shoulder scales with a specially designed identity synchronization regularisation.  We have shown the merits of the proposed approach through experimentation on SYSU-MM01 and RegDB datasets, and extensive ablative analysis have been conducted to validate our model design rationale.

\section{Acknowledgment}
This work is supported by the National Natural Science Foundation of China (No. 62173302).
\ifCLASSOPTIONcaptionsoff
  \newpage
\fi

\bibliography{IEEEfull}
\end{document}